\documentclass{article}

\usepackage{microtype}
\usepackage{graphicx}
\usepackage{subfigure}
\usepackage{amsmath,amsthm}
\usepackage{amsfonts}
\usepackage{bbm}
\usepackage{tikz}
\usetikzlibrary{bayesnet}
\usetikzlibrary{arrows,shadows,shapes,backgrounds,decorations,snakes,fit}
\usepackage{amssymb}

\let\emptyset\varnothing

\usepackage{paralist}
\usepackage{float}
\usepackage{booktabs} 
\usepackage{paralist, tabularx}
\usepackage{wrapfig}
\usepackage{enumitem}

\usepackage{hyperref}

\usepackage{xspace}
\newcommand{\name}[0]{VAEM\xspace}
\usepackage[final,nonatbib]{neurips_2020}

\title{VAEM: a Deep Generative Model for \\ Heterogeneous Mixed Type Data}

\author{
Chao Ma \\
University of Cambridge\\
\texttt{cm905@cam.ac.uk} \\
\And
Sebastian Tschiatschek \\
University of Vienna \\
\texttt{sebastian@tschiatschek.net}\\
\And
Jos{\'e} Miguel Hern{\'a}ndez-Lobato \\
University of Cambridge\\
\texttt{jmh233@cam.ac.uk} \\
\And
Richard Turner \\
University of Cambridge\\
\texttt{ret26@cam.ac.uk} \\
\And
Cheng Zhang \\
Microsoft Research Cambridge \\
\texttt{cheng.zhang@microsoft.com} \\
}

\begin{document}

\maketitle

\begin{abstract}

Deep generative models often perform poorly in real-world applications due to the heterogeneity of natural data sets. Heterogeneity arises from data containing different types of features (categorical, ordinal, continuous, etc.) and  features of the same type having different marginal distributions. 
We propose
an extension of 
variational autoencoders (VAEs) called \name
to
handle such heterogeneous data.
\name is a deep generative model that is trained in a two stage manner such that the first stage provides a more uniform representation of the data to the second stage, thereby sidestepping the problems caused by heterogeneous data.
We provide extensions of \name to handle partially observed data, and demonstrate its performance in data generation, missing data prediction and sequential feature selection tasks. Our results show that \name broadens the range of real-world applications where deep generative models can be successfully deployed.
\end{abstract}

\section{Introduction}
\label{sec:intro}

Variational Autoencoders (VAEs) \cite{kingma2013auto} are highly flexible probabilistic models, making them promising tools for enabling automated decision making under uncertainty in real-life scenarios. They are typically applied in standard settings in which each data dimension has a similar type and similar statistical properties (e.g.~consider the pixels of an image).
However, many real-world datasets contain variables with different types. For instance, in healthcare applications, a patient record may contain demographic information such as nationality which is of categorical type, the age which is ordinal, the height which is continuous.

 Naively applying vanilla VAEs to such mixed type heterogeneous data can lead to unsatisfying results. The reason for this is that it requires the use of different likelihood functions (e.g. Gaussian likelihoods for real-valued variables and Bernoulli likelihoods for binary variables). 
In this case, the contribution that each likelihood makes to the training objective can be very different, leading to challenging optimization problems \cite{kendall2018multi} in which some data dimensions
may be poorly-modeled in favor of others.  Figure \ref{fig:small_pp} (c) shows an example in which a vanilla VAE fits some of the categorical variables, but performs poorly on the continuous ones.

To overcome the limitations of VAEs in this setting, we propose  \underline{V}ariational \underline{A}uto-\underline{e}ncoder for heterogeneous \underline{m}ixed type data (VAEM)
and study its performance for decision making in real-world applications. \name uses a hierarchy of latent variables which is fit in two stages. In the first stage, we learn one type-specific VAE for each dimension. These initial one-dimensional VAEs 
capture marginal 
distribution properties and provide a latent representation that is more homogeneous across dimensions. 
In the second stage, another VAE is used to capture dependencies among the one-dimensional latent representations from the first stage. 

Our main contributions are:  
\begin{figure*}[t]
\centering
\centering

 \subfigure[Ground Truth]{
   \includegraphics[width=0.3\textwidth]{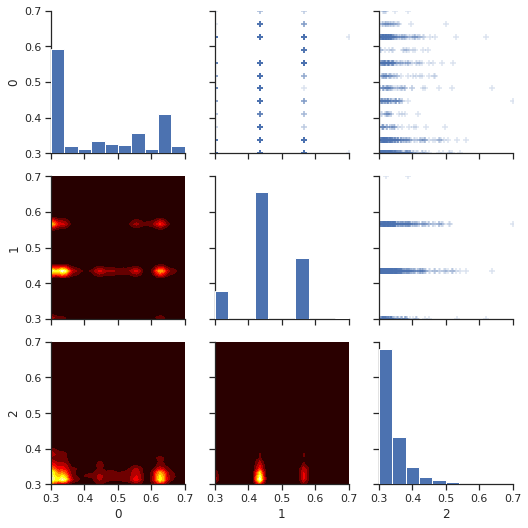}}
    \subfigure[\name(ours)]{\centering
  \includegraphics[width=0.3\textwidth]{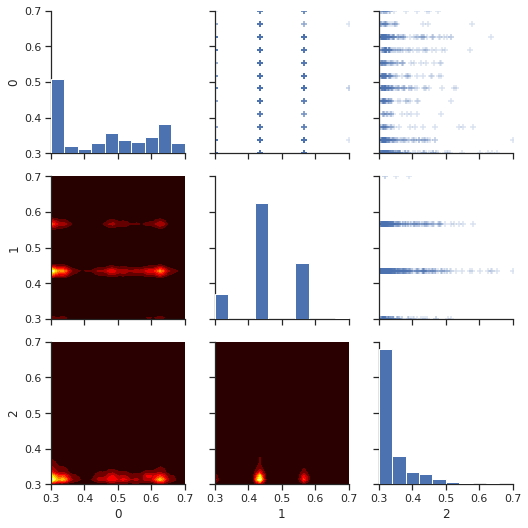}}
   \subfigure[VAE]{\centering
   \includegraphics[width=0.3\textwidth]{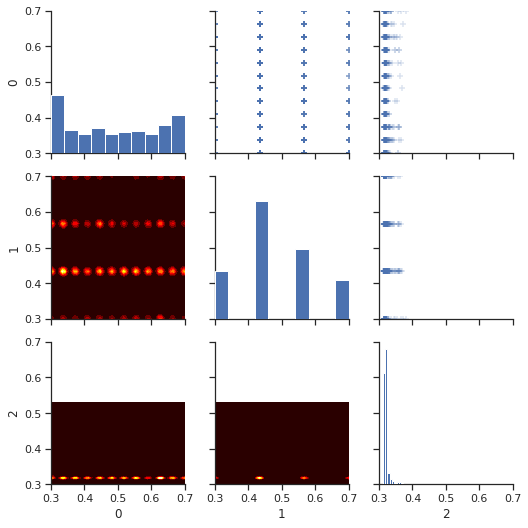}}
   
   \subfigure[VAE-extended]{
   \includegraphics[width=0.3\textwidth]{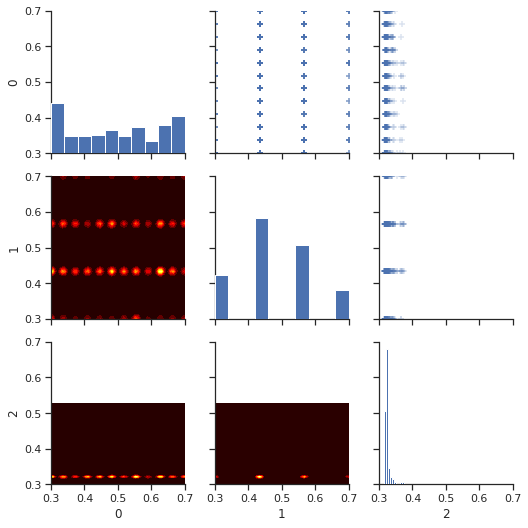}}
    \subfigure[VAE-balanced]{\centering
  \includegraphics[width=0.3\textwidth]{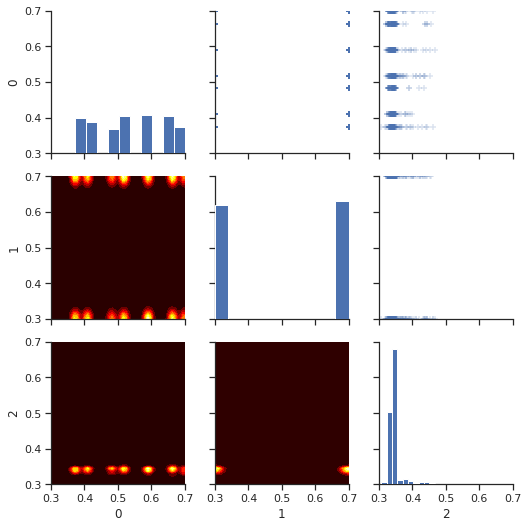}}
   \subfigure[VAE-HI]{\centering
   \includegraphics[width=0.3\textwidth]{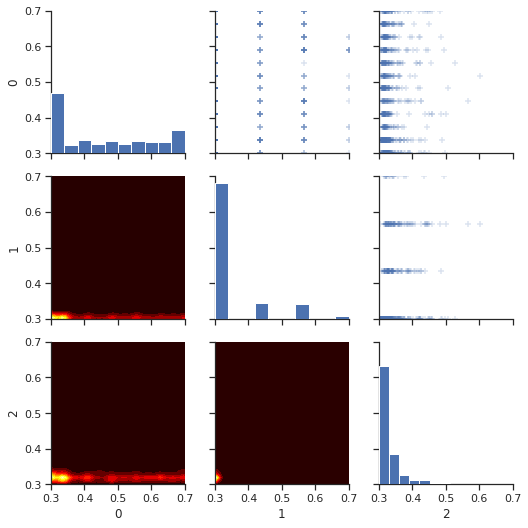}}
   
   \caption{Pair plots of 3-dimensional data generated from 5 different models (defined in Section \ref{sec:baseline_methods}) and  actual Bank data used to train them. 
    In each subfigure, diagonal plots show marginal histograms for each variable. 
   \name can correctly capture both continuous and discrete variables correctly both in terms of marginal distribution (diagonal plots) and pair-wise dependces (off-diagnal plots).}
   \label{fig:small_pp}
\end{figure*}

\begin{itemize}[nosep,leftmargin=1em,labelwidth=*,align=left]

    \item We present \name, a novel deep generative model for heterogeneous mixed-type data which alleviates the limitations of VAEs discussed above (See Section \ref{sec:method}). 
   
    We study the data generation quality of \name comparing with several existing VAE baselines on 5 different datasets. Our results show that \name can model mixed-type data more accurately than other baselines.
    \vspace{0.25cm}
    
    \item We extend \name to handle missing data and perform conditional data generation, and derive algorithms that enable it to be used for sequential active information acquisition (Section \ref{sec:SAIA}).
   
    We show that \name obtains strong performance for conditional data generation as well as sequential active information acquisition in cases where VAEs perform poorly. 
    
\end{itemize}
\section{VAE for heterogeneous mixed type data}
\label{sec:method}

In this section, we first review VAEs and their naive application to heterogeneous mixed-typed data. Then, we describe our proposed \name, 
a two stage model developed for such heterogeneous mixed type data, and the corresponding amortized inference method. Finally, we briefly discuss connections of VAEM with variational lower bounds and with data standardization methods.

\subsection{Background: variational auto-encoders}

Variational autoencoders (VAEs) \cite{kingma2013auto, rezende2014stochastic,zhang2018advances} employ deep generative latent variable models that are trained using amortized variational inference. As shown in Figure \ref{fig:tikz_VAE}, the VAE model assumes that the observed data $\mathbf{x}$ are generated from latent variable $\mathbf{z}$. The model is defined as 
$p_\theta(\mathbf{x}_n,\mathbf{z}_n) = p_{\mathbf{\theta}}(\mathbf{x}_n|\mathbf{z}_n) p(\mathbf{z}_n)$. 
Here, $p_{\mathbf{\theta}}(\mathbf{x}_n|\mathbf{z}_n)$ is often realized by a neural network known as the \emph{decoder}.
To approximate the posterior $p_\theta(\mathbf{z}_n | \mathbf{x}_n)$, VAEs use an encoder for \emph{amortized inference}, which takes the data $\mathbf{x}_n$ as input to produce the variational parameters of the approximate posterior $q_\phi(\mathbf{z}_n|\mathbf{x}_n)$. 
Finally, VAEs can be trained by optimizing the variational lower bound (ELBO).

With VAEs, the likelihood is typically fully factorized, thus $p(\mathbf{x}_n|\mathbf{z}_n) = \prod_d p_{d}({x}_{nd}|\mathbf{z}_n)$. In most machine learning applications, such as modeling images, each dimension of $\mathbf{x}_n$ has the same type, hence, each of these likelihood terms will take the same form, e.g.~Gaussian.

\textbf{VAE for mixed type data }~
A naive approach to handling heterogeneous mixed-typed data is to take the VAE model 
and use an appropriate likelihood function for each variable type.   
As discussed in Section \ref{sec:intro}, mixed likelihoods can cause problems in vanilla VAEs, causing them to perform poorly.

\subsection{VAE for heterogeneous mixed type data} \label{sec:MHM}

\begin{wrapfigure}{r}{0.4\textwidth}
\vspace{-30pt}
\centering
\resizebox{0.42 \textwidth}{!}{\subfigure[VAE]{
\resizebox{0.38 \linewidth}{!}{
\pgfdeclarelayer{background}
\pgfdeclarelayer{foreground}
\pgfsetlayers{background,main,foreground}

\begin{tikzpicture}

\tikzstyle{surround} = [thick,draw=black,rounded corners=1mm]
\tikzstyle{scalarnode} = [circle, draw, fill=white!11,  
    text width=1.2em, text badly centered, inner sep=2.5pt]
\tikzstyle{scalarnodenoline} = [  fill=white!11, 
    text width=1.2em, text badly centered, inner sep=2.5pt]
\tikzstyle{arrowline} = [draw,color=black, -latex]
\tikzstyle{dashedarrowcurve} = [draw,color=black, dashed, out=100,in=250, -latex]
\tikzstyle{dashedarrowline} = [draw,color=black, dashed,  -latex]

    \node [scalarnodenoline] at (1.7,0) (O)   {$\theta$};
    \node [scalarnodenoline] at (-1.7,0) (P)   {$\phi$};
    \node [scalarnode] at (0, 0) (Z) {$\mathbf{z}_n$};
    \node [scalarnode, fill=black!30, label=below right:D,  ] at (0, -1.5) (X) {$x_nd$};

    \node[surround, inner sep = .4cm,label=below right:N ] (f_N) [fit = (X), ] {};
    \node[surround, inner sep = .9cm, ] (f_N) [fit = (X)(Z) ] {};
    \path [arrowline] (Z) to (X);
    \path[arrowline]  (O) to (X);
    \path[dashedarrowline]  (P) to (Z);
    \path[dashedarrowcurve]  (-0.35,-1.5) to (-0.35,0);
\end{tikzpicture}}
\label{fig:tikz_VAE}
}
\subfigure[\name]{
\resizebox{0.5 \linewidth}{!}{








\pgfdeclarelayer{background}
\pgfdeclarelayer{foreground}
\pgfsetlayers{background,main,foreground}

\begin{tikzpicture}

\tikzstyle{surround} = [thick,draw=black,rounded corners=1mm]
\tikzstyle{scalarnode} = [circle, draw, fill=white!11,  
    text width=1.2em, text badly centered, inner sep=2.5pt]
\tikzstyle{scalarnodenoline} = [  fill=white!11, 
    text width=1.2em, text badly centered, inner sep=2.5pt]
\tikzstyle{arrowline} = [draw,color=black, -latex]
\tikzstyle{dashedarrowcurve} = [draw,color=black, dashed, out=100,in=250, -latex]
\tikzstyle{dashedarrowline} = [draw,color=black, dashed,  -latex]

    \node [scalarnodenoline] at (1.7,0) (O)   {$\theta$};
    \node [scalarnodenoline] at (-3.2,-1.0) (P)   {$\psi$};
    \node [scalarnodenoline] at (-3.2,-2.3) (A)   {$\phi$};
    \node [scalarnodenoline] at (-3.2,0) (Q)   {$\lambda$};
    \node [scalarnode] at (0, 0) (Z) {$\mathbf{h}_n$};
    \node [scalarnode, fill=black!30,label=below right:D,  ] at (0, -1.5) (X) {$x_{nd}$};
    \node [scalarnode,  ] at (-1.5,-1.5) (Z_loc)   {$z_{nd}$};

    \node[surround, label=below right:N,inner sep = .4cm, ] (f_N) [fit = (X)(Z_loc), ] {};
    
    \node[surround, inner sep = .9cm, ] (f_N) [fit = (X)(Z)(Z_loc)] {};
    \path [arrowline,color=red] (Z) to (Z_loc);
    \path[arrowline,color=blue]  (O) to (X);
    \path[arrowline,color=blue]  (Z_loc) to (X);
    \path[arrowline,color=red]  (P) to (Z_loc);
    \path[dashedarrowline,color=blue]  (A) to (Z_loc);
    \path[dashedarrowline,color=red]  (Q) to (Z);
    \path[dashedarrowline,color=red]  (-1.4,-1.1) to (-0.35,-0.05);
    \path[dashedarrowline,color=red]  (-0.0,-1.1) to (-0.0,-0.4);
    \path[dashedarrowline,color=blue]  (-0.35,-1.7) to (-1.1,-1.7);
\end{tikzpicture}}
\label{fig:tikz_MHMVAE}
}}
\caption{Graphical representations of the vanilla VAE and our VAEM where solid arrows denote decoders, and dashed arrows are encoders.
} 
\label{fig:tikz_all}
\end{wrapfigure}
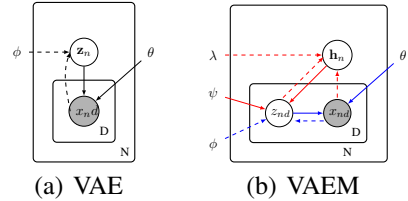

In order to properly handle mixed type data with heterogeneous marginals, our proposed method fits the data in a two-stage process. As shown in Figure \ref{fig:tikz_MHMVAE}, in the first stage we fit a different VAE independently to each data dimension $x_{nd}$. We call the resulting $D$ models \emph{marginal VAEs}. Then, in the second stage, in order to capture the inter-variable dependencies, a new multi-dimensional VAE, called the \emph{dependency network}, is build on top of the latent representations provided by the first-stage encoders. $D$ denotes the dimension of the observations and $N$ the number of data points with $x_{nd}$ being the $d$th dimension of the $n$th point.  We present the details below. 

\textbf{\emph{Stage one: training individual marginal VAEs to each single variable.}} In the first stage,  we focus on modeling the marginal distributions of each variable by training $D$ individual VAEs $p_{\mathbf{\theta}_d}(x_{nd}) =\mathbb{E}_{p(z_{nd})} p_{\mathbf{\theta}_d}(x_{nd}|z_{nd}),\ \ \forall{d} \in \{1,2,...,D\} $ independently, i.e.~each one is trained to fit a single dimension $x_{nd}$ from the dataset:
\begin{align}
     (\theta_d^\star,\phi_d^\star) =  {\arg \max}_{\theta_d,\phi_d}
     \sum_n \mathbb{E}_{q_{\phi_d}(z_{nd}|x_{nd})} \log\frac{ p_{\theta_d}(x_{nd},z_{nd})}{q_{\phi_d}(z_{nd}|x_{nd})}
\ \forall{d} \in \{1,2,...,D\}, \label{eq:Stage_one}
\end{align} where $p(z_{nd})$ is the standard Gaussian prior and $q_{\mathbf{\phi}_d}(z_{nd}|x_{nd})$ is the Gaussian encoder of the $d$-th marginal VAE. To specify the likelihood terms $p_{\theta_d}(x_{nd}|z_{nd})$, we use Gaussian likelihoods for continuous data and categorical likelihoods with one-hot representation for categorical data. The case of other variable types is discussed in Appendix \ref{app:exp_set_model}. 

Note that {Equation} \ref{eq:Stage_one} contains $D$ independent objectives. Each VAE $p_d(x_{nd}; \mathbf{\theta}_d)$ is trained independently and is only responsible for modeling the individual statistical properties of a single dimension $x_{nd}$ from the dataset. Thus, we assume that $z_{nd}$ is a scalar without loss of generality, although it would be trivial to use a multi-dimensional $\mathbf{z}_{nd}$ instead.%
Each marginal VAE can be trained independently until convergence \cite{dai2019diagnosing}, hence avoiding the optimization issues of vanilla VAEs. 
We then fix the  parameters of each marginal VAEs to be $\theta_d^\star$. 
These marginal VAEs fit the data well in practice, as shown in Figure \ref{fig:small_pp}, and have
tight ELBOs, as discussed in Appendix \ref{app:approximation_errors_of_marginal_VAEs}.

 \textbf{\emph{Stage two: training a dependency network to connect the marginal VAEs.}}
 In the second stage, we train a new multi-dimensional VAE on 
 top of the latent representations $\mathbf{z}$ provided by the encoders of the first-stage marginal VAEs.
 This additional VAE, 
 $p_{\psi}(\mathbf{z}) = \mathbb{E}_{p(\mathbf{h})} p_{\psi}(\mathbf{z}|\mathbf{h})$,
 models the inter-variable statistical dependencies 
  and is called the \emph{dependency network}.
  Here, $\mathbf{h}$ are the latent variables for the dependency network. Specifically, we train $p_{\psi}(\mathbf{z})$ as follows:
 \begin{align}
     \mathbf{x}_{n} &\sim p_{\text{data}}(\mathbf{x}) \label{eq:x}, z_{nd} \sim q_{\mathbf{\phi}_d}(z_d|x_{nd}), \ \ \forall{d} \in \{1,...,D\}, \\
      (\psi^\star,\lambda^\star) &\propto  \arg \max_{(\psi,\lambda)} \sum_n \mathbb{E}_{q_{\lambda}(\mathbf{h}_n|\mathbf{z}_n,\mathbf{x}_{n})} \log\frac{ p_{\psi}(\mathbf{z}_n,\mathbf{h}_n)}{q_\lambda(\mathbf{h}_n|\mathbf{z}_n,\mathbf{x}_{n})}.
      \label{eq:stage_two}
 \end{align}
The above procedure effectively disentangles the heterogeneous marginal properties of mixed type data (modelled by  the marginal VAEs), from the inter-variable dependencies (modelled by the dependency network). We call our model \emph{VAE for heterogeneous mixed type data (VAEM)}. After training the marginal VAEs and dependency network, our final generative model is given by
\begin{equation}
p_\theta(\mathbf{x}) = \mathbb{E}_{(\mathbf{z},\mathbf{h}) \sim p(\mathbf{h})\prod_d p_{\psi}(z_d|\mathbf{h})} \left[\prod_d p_{{\theta}_d}(x_d|z_d)\right]. \label{eq:final_mode}
\end{equation}
To handle complicated statistical dependencies,
we use the VampPrior \cite{tomczak2017vae}, which specifies a mixture of Gaussians (MoGs) as the prior distribution   for the high-level latent variables, 
 i.e., $ p(\mathbf{h}) = \frac{1}{K}\sum_k q_\lambda(\mathbf{h}|\mathbf{u_k})$, where
$K \ll N$ and the $\{\mathbf{u_k}\}$ are a subset of data points. 

\subsection{Discussions}

\textbf{Optimization objective} and relation to VAEs~
In Appendix \ref{app:proof_vaem}, we prove that the two stages of VAEM optimize a variational lower bound on the model likelihood
{$ \sum_n  \log \mathbb{E}_{p_\psi(\mathbf{z}_n)}  \prod_d  p_{\theta_d}(x_{nd}|z_{nd})  $}. In the first stage, we initialize $p_{\psi}(\mathbf{z}_n)$ to be a fully factorized standard Gaussian $p(\mathbf{z}_n)$ and keep it fixed, and optimize the rest of the parameters. This is obviously not an accurate prior since it does not consider dependencies among features. Thus, in the second stage, VAEM captures these dependencies by optimizing $p_{\psi}(\mathbf{z}_n)$ using the dependence VAE with variational distribution $q_{\lambda}(\mathbf{h}_n|\mathbf{z}_n,\mathbf{x}_n)$. 

\textbf{VAEM as advanced data standardization}~
In \name, the latent representations $z_d$ provided by the marginal VAEs are ``standardized'' in the sense that they have uniform
properties across dimensions. Each of these variables is encouraged to be close to a standard Gaussian thanks to the regularization effect from the Gaussian prior $p(z_{nd})$. In this way, we sidestep the heterogeneous mixed-type problem and the dependency VAE can focus on dependencies among homogeneous representations.


\section{\name for sequential active information acquisition}
\label{sec:SAIA}

The sequential active information acquisition task, described by \cite{ma2018eddi}, is an important application
to evaluate the capability of generative models for decision making under uncertainty.
To apply \name to this task, we
extend \name to handle missing data and estimate the Lindley information \cite{lindley1956measure}.

\subsection{Problem formulation} \label{sec:SAIA}

Suppose that, for a data instance $\mathbf{x} $, $\mathbf{x}_O$ denotes the set of currently observed variables, and $\mathbf{x}_U$ the unobserved ones. We are interested in predicting a target variable $\mathbf{x}_\Phi \in \mathbf{x}_U$ of interest given corresponding observed features $\mathbf{x}_O$ ($\mathbf{x}_\Phi \cap \mathbf{x}_O = \emptyset$).  
In this setting, a key problem is \emph{sequential active information acquisition} (SAIA): how to decide which variable $\mathbf{x}_i \subset \mathbf{x}_{U \setminus \Phi}$ to observe next, so that we optimally increase our knowledge (e.g., predictive ability) regarding $\mathbf{x}_\Phi$?

 As discussed in \cite{ma2018eddi}, to solve this problem we must have: 1) a good generative model that can handle missing data, and which can effectively generate conditional samples from $ p(\mathbf{x}_U|\mathbf{x}_O)$, 2) the ability to estimate a reward function given by the Lindley information. We now present extensions of VAEM to fulfill these two requirements.

\begin{wrapfigure}{r}{0.4\textwidth}
\centering
\vspace{-20pt}
\resizebox{0.40 \textwidth}{!}{
\pgfdeclarelayer{background}
\pgfdeclarelayer{foreground}
\pgfsetlayers{background,main,foreground}

\definecolor{babyblue}{rgb}{0.1, 0.6, 0.9}
\definecolor{bisque}{rgb}{1.0, 0.5, 0.8}
\definecolor{bittersweet}{rgb}{1.0, 0.8, 0.8}

\begin{tikzpicture}

\tikzstyle{surround} = [thick,draw=black,rounded corners=1mm]
\tikzstyle{scalarnode} = [circle, draw, fill=white!11,  
    text width=1.2em, text badly centered, inner sep=2.5pt]
\tikzstyle{scalarnodenoline} = [  fill=white!11, 
    text width=1.2em, text badly centered, inner sep=2.5pt]
\tikzstyle{arrowline} = [draw,color=black, -latex]
\tikzstyle{dashedarrowcurve} = [draw,color=black, dashed, out=100,in=250, -latex]
\tikzstyle{dashedarrowline} = [draw,color=black, dashed,  -latex]

\node [] at (0.2+8,0.5) (ip2) {\small $(|O|, M)$};

\node (rect) at (0+8,0) [draw,thick, fill=babyblue, minimum width=1.5cm, minimum height=0.5cm] (e1){$e_1$};

\node [] at (0+8,-1) {$...$};

\node (rect) at (0+8,-1.5) [draw,thick,fill=babyblue,minimum width=1.5cm, minimum height=0.5cm] (e4) {$e_{|O|}$};

\node (rect) at (0.7+8,-0.7) [draw,thick, fill=bisque, minimum width=0.75cm, minimum height=0.5cm] (x1){$v_1$};

\node (rect) at (0.7+8,-2.2) [draw,thick,fill=bisque,minimum width=0.75cm, minimum height=0.5cm] (x4){$v_{|\mathcal{V}|}$};

\node [circle] at (1.2+8+0.2,-0.3) [draw,thick, fill=black!20, minimum width=0.3cm, minimum height=0.3cm,inner sep=0pt] (m1){$\times$};

\node [circle] at (1.2+8+0.2,-1.8) [draw,thick,fill=black!20,minimum width=0.3cm, minimum height=0.3cm, inner sep=0pt] (m4){$\times$};

\node (rect)  at (2.5+8+0.2,-0.5) [draw, rounded rectangle,thick,inner sep=0pt, fill=green!50, minimum width=1.5cm, minimum height=0.35cm] (h1) { \small $l(\cdot)$};

\node [scalarnodenoline] at (2.25+8+0.2,-1) (s1) {shared};

\node (rect) at (2.5+8+0.2,-1.6) [draw, rounded rectangle,thick,inner sep=0pt,fill=green!50,minimum width=1.5cm, minimum height=0.35cm] (h4){  \small $l(\cdot)$};

\node (circle) at (4+8,-1) [draw,thick, fill=black!20, minimum width=0.4cm, minimum height=0.4cm] (g2) {$g(\cdot)$};
\path [arrowline]  (x1) to (m1);
\path [arrowline]  (x4) to (m4);
\path [arrowline]  (e1) to (m1);
\path [arrowline]  (e4) to (m4);
\path [arrowline]  (m1) to (h1);
\path [arrowline]  (m4) to (h4);
\path [arrowline]  (h1) to (g2);
\path [arrowline]  (h4) to (g2);

\node [] at (5+8,-0.5) (ip) {\small $(1,K)$};

\node (rect) at ( 5+8,-1) [draw,thick, fill=bittersweet, minimum width=0.8cm, minimum height=0.4cm] (c2) {$ c$};

\path [arrowline]  (g2) to (c2);

\node (rect) at ( 6+8,-1) [draw,thick, rotate=90, fill=black!20, minimum width=0.4cm, minimum height=0.4cm] (conc2) {Mean and Variance};

\path [arrowline]  (c2) to (conc2);

\node (rect) at (2.4+8,-0.9) [draw,thick,dashed,minimum width=6.5cm, minimum height=3.3cm] {};

\end{tikzpicture} }
\caption{Illustration of the partial inference net for our dependency network. 
}
\label{fig:tikz_PN}
\end{wrapfigure}

\subsection{Partial dependency network for handling missing data} \label{sec:partial}

The amortized inference network of \name (Section \ref{sec:MHM}) cannot handle partially observed data, since the number of observed variables $\mathbf{x}_O$ might vary across different data instances. Inspired by the Partial VAE \cite{ma2018eddi}, we apply a PointNet to build a partial inference network in the dependency VAE that infers $\mathbf{h}$ from partial observations. During the first stage, we estimate each marginal VAE with only the observed samples for that dimension: 
\begin{align}
     (\theta_d^\star,\phi_d^\star) =  \arg \max_{\theta_d,\phi_d} \sum_n  \mathbbm{1}_{ \{  \mathbf{x}_{n,O} \} } (x_{nd})
\mathbb{E}_{q_{\phi_d}(z_{nd}|x_{nd})} 
\log\frac{ p_{\theta_d}(x_{nd},z_{nd})}{q_{\phi_d}(z_{nd}|x_{nd})}, \ \ \forall{d} \in \{1,2,...,D\},\nonumber \label{eq:Stage_one_partial}
\end{align}where $\mathbbm{1}_{ \{  \mathbf{x}_{n,O} \} } (x_{nd})$ takes value one iff $x_{nd} \in \mathbf{x}_{n,O}$ and zero otherwise, with $\mathbf{x}_{n,O}$ being the set of observed variables for the $n$-th data instance.

 For the second stage, we need a dependency VAE that can handle partial observations.
Similarly as in the partial-VAE \cite{ma2018eddi}, in the presence of missing data, the dependency VAE specifies $p_{\mathbf{z}_O}(\mathbf{z}_O;\psi) = \mathbb{E}_{p(\mathbf{h})} \prod_{d\in O}p_{\psi}(z_d|\mathbf{h})$. This dependency VAE is trained by maximizing the partial ELBO:
 \begin{align}
\mathbb{E}_{q_{\lambda}(\mathbf{h}|\mathbf{z}_O,\mathbf{x}_O)} \log\frac{ \prod_{d\in O}p_{\psi}(z_d,\mathbf{h})p(\mathbf{h})}{q_\lambda(\mathbf{h}|\mathbf{z}_O,\mathbf{x}_O)}, 
      z_d &\sim q_d(z_d|x_{{\text{data}},d},\mathbf{\phi}_d) \ \ \forall{d} \in O, \ \  \mathbf{z}_O  = \{z_d|d \in O \} 
 \end{align}

 where $\mathbf{h}$ is the latent variable of the dependency network, $q_\lambda(\mathbf{h}|\mathbf{z}_O,\mathbf{x}_O)$ is a set-function, the so-called \emph{partial inference net}, the structure of which is shown in Figure \ref{fig:tikz_PN}. Essentially, for each feature in $\mathbf{x}_O$, the input to the partial inference net is first modified as
 {$\mathbf{s}_O := \{ v \times \mathbf{e}_v|v \in \mathbf{z}_O \cup \mathbf{x}_O\}$} using element-wise multiplication, and $\mathbf{e}_v$ is a \emph{feature embedding}\footnote{If $v$ is a non-continuous variable such as categorical,  the operation $v\times\mathbf{e}_v$ is performed on the one-hot representation of $v$, as detailed in Appendix \ref{app:exp_set_model}}.  $\mathbf{s}_O$ is then fed into a \emph{feature map} (a neural network) $l(\cdot): \mathbb{R}^{M} \rightarrow \mathbb{R}^{K}$, where $M$ and $K$ is the dimension of the feature embedding and the feature map, respectively. Finally, we apply a permutation invariant aggregation operation $g(\cdot)$, such as summation.  In this way, $q_\lambda(\mathbf{h}|\mathbf{z}_O,\mathbf{x}_O)$ is invariant to the permutations of the elements of $\mathbf{x}_O$, and $\mathbf{x}_O$ can have arbitrary length.

\paragraph{Approximate conditional data generation} Once the marginal VAEs and the partial dependency network are trained, we can generate conditional samples that approximate $p_{\theta}(\mathbf{x}_U|\mathbf{x}_O)$ by the following inference procedure: first, the latent representations $z_d$ for the observed variables are inferred. With this representation, we use the partial inference network to infer $\mathbf{h}$, which is the latent code for the second stage VAE. Given $\mathbf{h}$, we can generate the $z_s$ which are the latent code for the unobserved dimensions and then generate the $x_s$: 
\begin{align}
     &z_d \sim q_d(z_d|x_{d},\mathbf{\phi}_d) \ \ \forall{d} \in O, \ \  \mathbf{z}_O  = \{z_d|d \in O \},
     \mathbf{h} \sim q_{\lambda}(\mathbf{h}|\mathbf{z}_O,\mathbf{x}_O),
     z_s \sim p_{\psi}(z_s|\mathbf{h}) \ \ \forall{s} \in U, \ \ \nonumber \\
    & \mathbf{z}_U  = \{z_s|s \in U \}, 
     x_s \sim p_{\theta}(x_s|z_s) \ \ \forall{s} \in U, \ \  \mathbf{x}_U  = \{x_s|s \in U \}. \label{eq:con_gen}
 \end{align}{}
\subsection{Reward estimation with VAEM}

Following \cite{ma2018eddi}, SAIA can be framed as a Bayesian experimental design problem. The next variable to observe, $\mathbf{x}_i \subset \mathbf{x}_{U \setminus \Phi}$, is selected by the following \emph{information reward} function:
\begin{equation*} \label{eq:IR}
R_I(\mathbf{x}_i,\mathbf{x}_O) = \mathbb{E}_{\mathbf{x}_i \sim p(\mathbf{x}_i|\mathbf{x}_O)}
\mathbb{KL}\left[p(\mathbf{x}_\Phi | \mathbf{x}_i,\mathbf{x}_O) \,\|\, p(\mathbf{x}_\Phi | \mathbf{x}_O)
\right].
\end{equation*}
Where $\mathbb{KL}$ is the Kullback-Leibler divergence. Intuitively this reward function selects a variable if it can result in the most drastic change in our current belief on $\mathbf{x}_\Phi$. Such change is captured by $\mathbb{KL}\left[p(\mathbf{x}_\Phi | \mathbf{x}_i,\mathbf{x}_O) \,\|\, p(\mathbf{x}_\Phi | \mathbf{x}_O)
\right]$.

We use a trained partial \name model (\ref{eq:con_gen}) to estimate the required distributions $p(\mathbf{x}_i|\mathbf{x}_O)$, $p(\mathbf{x}_\Phi | \mathbf{x}_i,\mathbf{x}_O)$, and  $p(\mathbf{x}_\Phi | \mathbf{x}_O)$. Due to the intractability of $\mathbb{KL}\left[p(\mathbf{x}_\Phi | \mathbf{x}_i,\mathbf{x}_O) \,\|\, p(\mathbf{x}_\Phi | \mathbf{x}_O)
\right]$, we must resort to approximations. An efficient estimation of $R_I(\mathbf{x}_i,\mathbf{x}_O)$ was proposed by \cite{ma2018eddi}, where the computations are reduced to a series of KL divergences in latent space. However, it cannot be applied in our case since our model is hierarchical and contains latent variables $\{ \mathbf{z}_U \}$ with variable size due to missingness from $\mathbf{x}_U$. We hereby extend their method and show that $R_I(\mathbf{x}_i,\mathbf{x}_O)$ can be approximated as follows (Appendix \ref{sec:app_information}): 
\begin{align} \label{eq:reward_approx}
\hat{R}_I(\mathbf{x}_i,\mathbf{x}_O) & = \mathbb{E} _{  p_\theta(\mathbf{x}_i,\mathbf{z}_i,\mathbf{z}_O|\mathbf{x}_O)} \left\{ \mathbb{KL}\left[ q_\lambda(\mathbf{h}|\mathbf{z}_i,\mathbf{z}_O)||q_\lambda( \mathbf{h}|\mathbf{z}_O)\right]\right. -\\ \nonumber
 &\quad\left. \mathbb{E} _{p_\theta(\mathbf{x}_{\phi},\mathbf{z}_\Phi, |\mathbf{x}_O)}  \mathbb{KL}\left[ q_\lambda(\mathbf{h}|\mathbf{z}_{\Phi},\mathbf{z}_i,\mathbf{z}_O)||q_\lambda( \mathbf{h}|\mathbf{z}_{\Phi}, \mathbf{z}_O) \right] \right\}.
\end{align}
Note that, for compactness, we omitted the inputs $\mathbf{x}_O$ and $\mathbf{x}_i$ to the partial inference networks. The approximation (\ref{eq:reward_approx}) is very efficient to compute, since all KL terms can be calculated analytically, assuming that the partial inference net $q_\lambda(\mathbf{h}|\mathbf{z}_O)$ is Gaussian (or other common distributions where KL divergences can be estimated deterministically or via Monte Carlo).

\subsection{Enhancing the predictive performance of \name}
\label{sec:predictive}
To predict the variable of interest
is desirable to use a supervised learning method instead of just an unsupervised method such as the VAE. In active information acquisition, the target of interest $\mathbf{x}_\Phi$ ($\mathbf{x}_\Phi \cap \mathbf{x}_O = \emptyset$) is often the variable that we try to predict in a regression/classification task. In order to enhance the predictive performance of \name, we propose to use the following factorization: 
\begin{align}
    p_\theta(\mathbf{x}_O, \mathbf{x}_\Phi)  =
\mathbb{E}_{p_\theta(\mathbf{x}_{U\setminus \Phi},\mathbf{h}|\mathbf{x}_O)} p_\gamma(\mathbf{x}_\Phi|\mathbf{x}_O, \mathbf{x}_{U\setminus \Phi}, \mathbf{h}) p_\theta(\mathbf{x}_O),
\label{eq:predictive}
\end{align}
where $p_\gamma(\mathbf{x}_\Phi|\mathbf{x}_O, \mathbf{x}_{U\setminus \Phi},\mathbf{h})$ is the discriminator (prediction model) that takes the observed variables $\mathbf{x}_O$, the imputed variables $\mathbf{x}_{U\setminus \Phi}$ and the global latent representation $\mathbf{h}$ as input, and predicts the distribution of $\mathbf{x}_\Phi$. Training the joint model in  Equation ~\ref{eq:predictive} is similar to our previous two-stage procedure, which is detailed in Appendix \ref{sec:app_predictive}.

\section{Related works}
\name is a two stage method that extends generative models to handle mixed type heterogeneous data, with applications in  SAIA. Therefore, we review here the literature in the following aspects.

\textbf{Generative models for mixed-type heterogeneous data}~ This type of models are under-explored in the literature. \cite{nazabal2018handling} proposed Heterogeneous-Incomplete VAE (HI-VAE), a deep generative model with heterogeneous variables. It uses a multi-head decoder architecture \cite{nguyen2017variational, gordon2018meta}. However, this does not help balance the learning of different marginal distributions. A recent empirical study \cite{ma2019hm} shows that HI-VAE fails to recover the marginal distributions correctly. Finally, another orthogonal line of work focuses on using traditional latent variable models to infer the variable types automatically \cite{dhirautomatic,dhirautomatic2,valera2017automatic, hernandez2014learning}. However, they have been shown to be surpassed by VAE-based models empirically \cite{nazabal2018handling}.

\textbf{VAEs for Sequential Active Information Acquisition}~ There has been a recent drive to develop methods for sequential active information acquisition (SAIA) \cite{ma2018eddi,gong2019icebreaker, saar2009active, thahir2012efficient}, where the goal is to optimally acquire one variable at a time for each data instance. SAIA differs from traditional active learning (AL) \cite{lindley1956measure, mackay1992information, settles2012active}

settings in that AL performs \emph{instance-wise} selection for best predictive performance, whilst SAIA  performs \emph{variable-wise acquisition.}
SAIA tasks are usually solved by partial VAEs \cite{ma2018eddi,gong2019icebreaker}, which has been shown to be successful in continuous variables settings. Our work further demonstrates how to use VAEM to solve SAIA in the mixed type data setting.

\textbf{Two-stage methods for VAEs}~ An orthogonal line of work (TS-VAE) \cite{dai2019diagnosing}, uses a two-stage training paradigm to improve VAEs when applied to homogeneous continuous data. First, a VAE is trained on the data to discover low-dimensional latent representations; Then, another VAE is trained on these representations to capture any deviations from the prior in the latent space of the first VAE. In this regard, this method is similar to the VampPrior VAE \cite{tomczak2017vae}, where a mixture of Gaussians is optimized to provide a flexible prior on the VAE latent variables. However, unlike our approach, both TS-VAE and VampPrior do not naturally handle mixed type data and missing data.

\section{Experiments}

In this section, we evaluate the performance and validity of \name. We focus on three different tasks with mixed type heterogeneous data: 1) data modeling and generation, 2) conditional data generation (imputation) and 3) sequential active information acquisition.

\begin{table}[h]
\setlength{\tabcolsep}{2pt}
\begin{minipage}[b]{0.50\linewidth}

\centering
\caption{Data generation quality in terms of average test NLL per variable and corresponding standard errors.}
\resizebox{1.0\linewidth}{!}{
\begin{tabular}{p{0.6in}p{0.57in}p{0.57in}p{0.57in}p{0.57in}p{0.57in}p{0.57in}} \toprule
\textbf{Method} & VAEM (Ours) & VAE & VAE-balanced & VAE-extended & VAE-HI \\ \midrule
Bank & \textbf{-1.15$\pm$.09}& \ 2.09$\pm$.04 & \ 0.72$\pm$.01 & \ 2.06$\pm$.00 & -0.72$\pm$.00\\ 
Boston & \textbf{-2.16$\pm$.01} & -1.69$\pm$.01 & \ 0.38$\pm$.01 & -1.61$\pm$.02 &\ 2.11$\pm$.01 \\
Avocado & \textbf{-0.16$\pm$.00}& \ 0.04$\pm$.00 & \ 1.32$\pm$.01& \ 0.04$\pm$.00 &\ 0.04$\pm$.00\\ 
Energy & -1.28$\pm$.09 & \textbf{-1.47$\pm$.07}& \ 0.69$\pm$.02& -1.46$\pm$.08 &\ 0.16$\pm$.00\\ 
MIMIC & \textbf{-1.01$\pm$.00}& \ 0.08$\pm$.00& \ 0.69$\pm$.00& \ 0.08$\pm$.00 & \ 0.08$\pm$.00\\
\bottomrule
Avg. Rank & \textbf{\ 1.40$\pm$.36}& \ 2.60$\pm$.61& \ 4.40$\pm$.36& \ 3.00$\pm$.40& \ 3.00$\pm$.57\\
\bottomrule
\end{tabular}
}
\label{tab:marginal_llh}

\end{minipage}\hspace{2pt}
\begin{minipage}[b]{0.50\linewidth}
\caption{Conditional data generation quality under random missing entries. 
Average test NLL per variable and corresponding standard errors.}
\resizebox{1.0\linewidth}{!}{
\begin{tabular}{p{0.6in}p{0.57in}p{0.57in}p{0.57in}p{0.57in}p{0.57in}p{0.57in}} \toprule
\textbf{Method} & VAEM (Ours) & VAE & VAE-balanced & VAE-extended & VAE-HI \\ \midrule
Bank & \textbf{-1.21$\pm$.12} & \ 2.09$\pm$.00 & \ 0.68$\pm$.00 & \ 2.09$\pm$.00 & -0.83$\pm$.01\\ 
Boston & \textbf{-2.18$\pm$.03} & -1.66$\pm$.02 & \ 0.37$\pm$.00 & -1.67$\pm$.01 & \ 1.58$\pm$.01\\
Avocado & \textbf{-0.15$\pm$.00}& \ 0.04$\pm$.00 & \ 1.33$\pm$.00& \ 0.04$\pm$.00& \ 0.04$\pm$.00\\ 
Energy & -1.30$\pm$.05 & \textbf{-1.50$\pm$.06}& \ 0.67$\pm$.01& -1.50$\pm$.06& \ 0.13$\pm$.00\\ 
MIMIC & \textbf{-1.10$\pm$.00}& \ 0.08$\pm$.00& \ 0.57$\pm$.00& \ 0.08$\pm$.00 &\ 0.08$\pm$.00\\ 
\bottomrule
Avg. Rank & \textbf{\ 1.40$\pm$.36}& \ 2.60$\pm$.61& \ 4.40$\pm$.38& \ 2.30$\pm$.44& \ 3.00$\pm$.57\\
\bottomrule
\end{tabular}
}
\label{tab:conditional_llh}
\end{minipage}
\end{table}

\subsection{Baselines and datasets} \label{sec:baseline} \label{sec:baseline_methods}
In the experiments, we consider a number of baselines. Unless otherwise specified, all VAE baselines use the partial inference network and the discriminator specified in Section \ref{sec:partial} and \ref{sec:predictive}, respectively. Moreover, all baselines are equipped with a MoG priors (Section \ref{sec:MHM}). Our main baselines include:
\begin{itemize}[nosep,leftmargin=1em,labelwidth=*,align=left]
    \item Heterogeneous-Incomplete VAE  \cite{nazabal2018handling}. We match the dimensionality of latent variables to be the same as our \name. We denote this by \texttt{VAE-HI}.
    
    \item VAE: A vanilla VAE equipped with a VampPrior \cite{tomczak2017vae}. The number of latent dimensions is the same as in the second stage ($\mathbf{h}$) of VAEM. We denote this by \texttt{VAE}. 
    
    \item VAE with extended latent dimension: same as the \texttt{VAE}, but with the latent dimension increased to be the same as VAEM (sum of the dimensions of $\mathbf{h}$ and $\mathbf{z}$). We denote this by \texttt{VAE-extended}.
    
   \item VAE with balanced likelihoods. This baseline automatically equal the scale of each likelihood term of the different variable types, by multiplying each likelihood term with an {adaptive constant (Appendix \ref{app:exp_set_model})}. We denote this baseline by \texttt{VAE-balanced}.
\end{itemize}{}

We use the same collection of mixed type datasets in all tasks:

\begin{itemize}[nosep,leftmargin=1em,labelwidth=*,align=left]
    \item  Two  standard UCI benchmark datasets: Boston housing and energy efficiency \cite{Dua:2017};
    \item Two relatively large real-world datasets: Bank marketing;\cite{moro2011using} and Avocado sales prediction.
    \item A real-world medical dataset: MIMIC III \cite{johnson2016mimic}, the largest public medical dataset for intensive care. We focus on the mortality prediction task. 
\end{itemize}{}
Details including model details, hyperparameters and data processing can be found in Appendix \ref{app:exp_set}.

\subsection{Mixed type data generation 
}
\label{sec:generation}

\begin{figure*}[t]
    \subfigure[Avocado]{
   \includegraphics[width=0.30\textwidth]{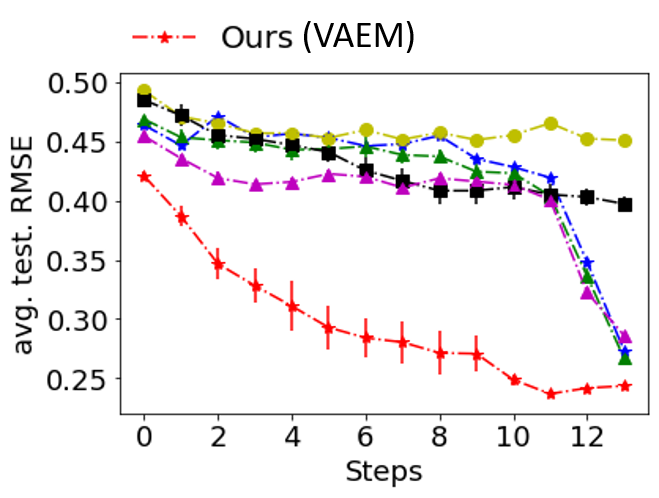}}
   \subfigure[Bank]{
   \includegraphics[width=0.31\textwidth]{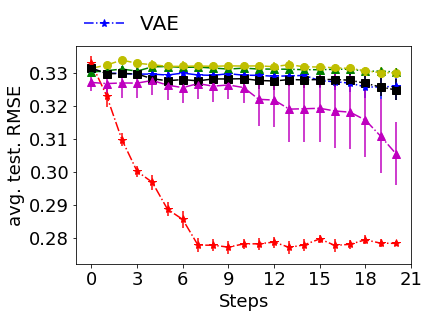}}
   \subfigure[MIMIC]{
   \includegraphics[width=0.30\textwidth]{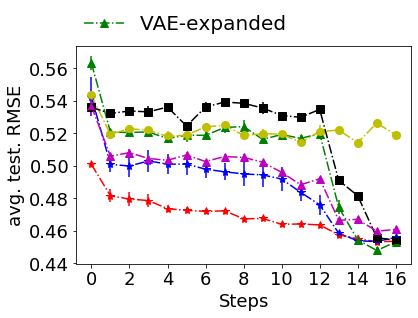}}
   \\
   \subfigure[Energy]{
   \includegraphics[width=0.30\textwidth]{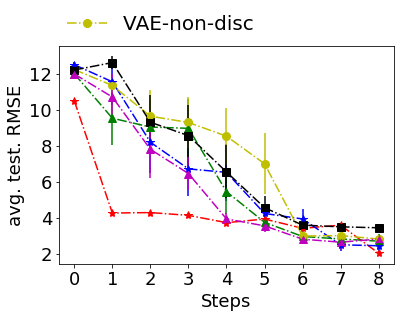}}
\subfigure[Boston]{
   \includegraphics[width=0.30\textwidth]{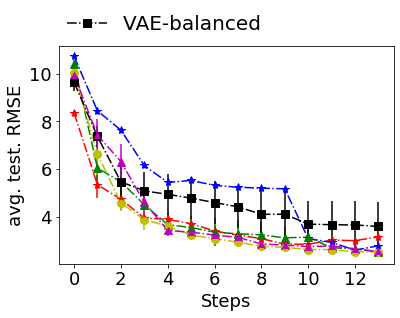}}
   \subfigure[AUIC Comparison]{
   \includegraphics[width=0.32\textwidth]{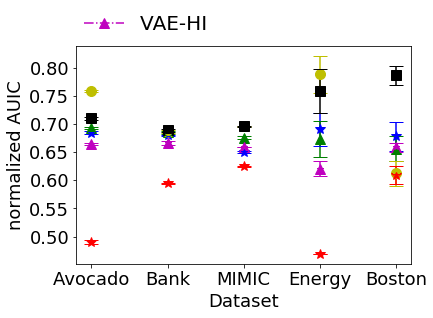}}
\caption{Information curves of sequential active information acquisition, with standard error as error bars. \textbf{((a)-(e))}: Information curves of Avocado sales, Bank marketing, MIMIC-III, Energy, Boston Housing, respectively. \textbf{(f)}: Comparison of AUIC on each dataset. All AUIC values are normalized by dividing by the average AUIC value within the corresponding dataset.   }
\vspace{-12pt}
   \label{fig:al_meta}
\end{figure*}  

In this task, we evaluate the quality of our generative model in terms of mixed type data generation. 
For all datasets, we first train the models and then quantitatively compare their performance using a 90\%-10\% train-test split. All experiments are repeated 5 times with different random seeds.

\textbf{Visualization by pair plots}~
In deep generative models, the data generation quality is indicative of how well the model describes the data. Thus, we first visualize the data generated by each model on a representative dataset: Bank marketing. This dataset contains three different data types with drastically different marginals, which present a challenge for learning. We fit our models to the Bank data and then generate pair plots for three variables, $x_0$, $x_1$ and $x_2$ (the first two are categorical and the third one is continuous), selected from the data (see Figure \ref{fig:small_pp}). Full plots with all the other variables can be found in Appendix \ref{app:exp_plots}. In each subfigure of Figure \ref{fig:small_pp}, diagonal plots show the marginal histograms of each variable. The upper-triangular part shows sample scatter plots for each variable pair. The lower-triangular part shows heat maps identifying regions of high-probability density for each variable pair, as given by kernel density estimates.

By comparing the plots in the diagonals of Figure \ref{fig:small_pp} (a) and Figure \ref{fig:small_pp} (c), we notice that vanilla VAE is able to describe the marginal distribution of the second categorical variable. However, it fails to mimic the behaviour of the third variable. Note that this variable (Figure \ref{fig:small_pp} (a)), which corresponds to the ``duration" feature of the dataset, has a heavy tail behaviour, which is ignored by vanilla VAE.  On the other hand, although the VAE-balanced model and VAE-HI (Figures \ref{fig:small_pp} (e) and (f)) can partially describe this heavy-tail behaviour, they fail to model the marginal distribution of the second categorical variable well. Unlike the baselines, our \name model (Figure \ref{fig:small_pp} (b)) is able to accurately describe the marginals and joint distributions for both categorical and heavy-tailed continuous distributions.

\textbf{Quantitative evaluation on all datasets}~
To evaluate the data generation quality quantitatively, we compute the average negative log-likelihood (NLL) of the models on the test sets (detailed in Appendix \ref{app:exp_set_hyper}). Note that all NLL numbers are divided by the actual number of variables in each dataset. As shown in Table \ref{tab:marginal_llh}, \name can consistently provide a very good fit of the data, and on average significantly outperforms the other baselines.

\subsection{Mixed type conditional data generation 
} \label{sec:exp_conditional}
An important aspect of generative models is their ability to perform conditional data generation. That is, given a data instance, to infer the posterior distribution of unobserved variables $\mathbf{x}_U$ given observed $\mathbf{x}_O$. For all baselines evaluated in this task, we train the partial version of them (i.e., generative
model + partial inference net \cite{ma2018eddi}). We manually drops 50\% of the data entries from test set for imputation.

Since all inference methods are probabilistic, we report the average test NLLs on unobserved data, as opposed to the imputation error, which is more typically used in the literature. We also provide imputation error results in Appendix \ref{app:exp_results}as reference.

Results are summarized in Table \ref{tab:conditional_llh}, where all NLL values have been divided by the number of observed variables. We repeat our experiments for 5 runs. 
Note that the automatic balancing strategy \textbf{VAE-balanced} almost always deteriorates performance. By contrast, Table \ref{tab:conditional_llh} shows that our proposed method is very robust, yielding significant improvements over all baselines on 4 out of 5 datasets.

\subsection{Sequential active information acquisition (SAIA) 
} \label{exp:SAIA}

In our final experiments, we apply \name to the task of sequential active information acquisition (SAIA) based on the formulation described in Section \ref{sec:SAIA}.
We use this task as an example to showcase how VAEM can be used in decision making under uncertainty. 
We employ the same experiment pipeline as in \cite{ma2018eddi}.
The reward function of \name is estimated according to Section \ref{sec:SAIA}. We add an additional baseline, denoted by \texttt{VAE-no-disc}, where the prediction of the target is directly generated by the decoder without using a predictive model. By comparing to this baseline, we can show the importance of the discriminator described in Section \ref{sec:predictive}. The other settings are the same as described in Section \ref{sec:baseline}. All baseline methods use the information reward estimation method proposed in \cite{ma2018eddi}. All experiments are repeated ten times. 

Figure \ref{fig:al_meta} shows the average test RMSEs on $\mathbf{x}_{\Phi}$ for each variable selection step on all five datasets, where $\mathbf{x}_{\Phi}$ is the target variable. The y-axis shows the error of the prediction and the x-axis shows the number of features acquired so far. The curves in Figure \ref{fig:al_meta} are called information curves \cite{ma2018eddi,gong2019icebreaker}. The area under the information curve (AUIC) can be used to evaluate the performance of a method in SAIA. The smaller the area, the better the method. From Figure \ref{fig:al_meta}, we see that \name  performs consistently better than the other baselines. Note that, on the Bank marketing and Avocado sales datasets, a lot of heterogeneous variables are involved and other baselines fail to reduce the test RMSE quickly and \name outperforms them by a large margin. These experiments show that \name is able to acquire information efficiently on mixed type datasets.

\section{Conclusion}
We proposed \name, a novel two stage deep generative model 
 that can handle mixed type data with heterogeneous marginals and missing data. \name sidesteps the problems arising from fitting heterogeneous data directly. For this, VAEM uses a two-stage training procedure. Efficient amortized inference methods and extensions are proposed. Experiments yield promising results, indicating that \name is useful for real-world applications of deep generative models. In future works, we will study other real-life applications and further extent the \name method. 



\bibliographystyle{plain}
\bibliography{example_paper}

\begin{thebibliography}{10}

\bibitem{dai2019diagnosing}
Bin Dai and David Wipf.
\newblock Diagnosing and enhancing vae models.
\newblock {\em arXiv preprint arXiv:1903.05789}, 2019.

\bibitem{Dua:2017}
Dua Dheeru and Efi Karra~Taniskidou.
\newblock {UCI} machine learning repository, 2017.

\bibitem{dhirautomatic}
Neil Dhir, Davide Zilli, Tomasz Rudny, and Alessandra Tosi.
\newblock Automatic type inference with a nested latent variable model.

\bibitem{dhirautomatic2}
Neil Dhir, Davide Zilli, Tomasz Rudny, and Alessandra Tosi.
\newblock Automatic type inferential general latent feature model.

\bibitem{gong2019icebreaker}
Wenbo Gong, Sebastian Tschiatschek, Richard Turner, Sebastian Nowozin, and
  Jos{\'e}~Miguel Hern{\'a}ndez-Lobato.
\newblock Icebreaker: Element-wise active information acquisition with bayesian
  deep latent gaussian model.
\newblock {\em arXiv preprint arXiv:1908.04537}, 2019.

\bibitem{gordon2018meta}
Jonathan Gordon, John Bronskill, Matthias Bauer, Sebastian Nowozin, and
  Richard~E Turner.
\newblock Meta-learning probabilistic inference for prediction.
\newblock {\em arXiv preprint arXiv:1805.09921}, 2018.

\bibitem{harutyunyan2017multitask}
Hrayr Harutyunyan, Hrant Khachatrian, David~C Kale, and Aram Galstyan.
\newblock Multitask learning and benchmarking with clinical time series data.
\newblock {\em arXiv preprint arXiv:1703.07771}, 2017.

\bibitem{hernandez2014learning}
Jos{\'e}~Miguel Hern{\'a}ndez-Lobato, James~Robert Lloyd, Daniel
  Hern{\'a}ndez-Lobato, and Zoubin Ghahramani.
\newblock Learning the semantics of discrete random variables: Ordinal or
  categorical.
\newblock In {\em NIPS Workshop on Learning Semantics}, 2014.

\bibitem{johnson2016mimic}
Alistair~EW Johnson, Tom~J Pollard, Lu~Shen, H~Lehman Li-wei, Mengling Feng,
  Mohammad Ghassemi, Benjamin Moody, Peter Szolovits, Leo~Anthony Celi, and
  Roger~G Mark.
\newblock Mimic-iii, a freely accessible critical care database.
\newblock {\em Scientific Data}, 3:160035, 2016.

\bibitem{kendall2018multi}
Alex Kendall, Yarin Gal, and Roberto Cipolla.
\newblock Multi-task learning using uncertainty to weigh losses for scene
  geometry and semantics.
\newblock In {\em Proceedings of the IEEE conference on computer vision and
  pattern recognition}, pages 7482--7491, 2018.

\bibitem{adam2015}
Diederik~P. Kingma and Jimmy~Lei Ba.
\newblock Adam: a method for stochastic optimization.
\newblock In {\em International Conference on Learning Representations}, pages
  1--13, 2015.

\bibitem{kingma2013auto}
Diederik~P Kingma and Max Welling.
\newblock Auto-encoding variational bayes.
\newblock {\em arXiv preprint arXiv:1312.6114}, 2013.

\bibitem{lindley1956measure}
Dennis~V Lindley.
\newblock On a measure of the information provided by an experiment.
\newblock {\em The Annals of Mathematical Statistics}, pages 986--1005, 1956.

\bibitem{ma2019hm}
Chao Ma, Sebastian Tschiatschek, Yingzhen Li, Richard Turner, Jose~Miguel
  Hernandez-Lobato, and Cheng Zhang.
\newblock Hm-vaes: a deep generative model for real-valued data with
  heterogeneous marginals.
\newblock 2019.

\bibitem{ma2018eddi}
Chao Ma, Sebastian Tschiatschek, Konstantina Palla, Jose Miguel~Hernandez
  Lobato, Sebastian Nowozin, and Cheng Zhang.
\newblock Eddi: Efficient dynamic discovery of high-value information with
  partial vae.
\newblock {\em arXiv preprint arXiv:1809.11142}, 2018.

\bibitem{mackay1992information}
David~JC MacKay.
\newblock Information-based objective functions for active data selection.
\newblock {\em Neural computation}, 4(4):590--604, 1992.

\bibitem{moro2011using}
Sergio Moro, Raul Laureano, and Paulo Cortez.
\newblock Using data mining for bank direct marketing: An application of the
  crisp-dm methodology.
\newblock In {\em Proceedings of European Simulation and Modelling
  Conference-ESM'2011}, pages 117--121. EUROSIS-ETI, 2011.

\bibitem{nazabal2018handling}
Alfredo Nazabal, Pablo~M Olmos, Zoubin Ghahramani, and Isabel Valera.
\newblock Handling incomplete heterogeneous data using vaes.
\newblock {\em arXiv preprint arXiv:1807.03653}, 2018.

\bibitem{nguyen2017variational}
Cuong~V Nguyen, Yingzhen Li, Thang~D Bui, and Richard~E Turner.
\newblock Variational continual learning.
\newblock {\em arXiv preprint arXiv:1710.10628}, 2017.

\bibitem{paquet2012hierarchical}
Ulrich Paquet, Blaise Thomson, and Ole Winther.
\newblock A hierarchical model for ordinal matrix factorization.
\newblock {\em Statistics and Computing}, 22(4):945--957, 2012.

\bibitem{rezende2014stochastic}
Danilo~Jimenez Rezende, Shakir Mohamed, and Daan Wierstra.
\newblock Stochastic backpropagation and approximate inference in deep
  generative models.
\newblock {\em arXiv preprint arXiv:1401.4082}, 2014.

\bibitem{saar2009active}
Maytal Saar-Tsechansky, Prem Melville, and Foster Provost.
\newblock Active feature-value acquisition.
\newblock {\em Management Science}, 55(4):664--684, 2009.

\bibitem{settles2012active}
Burr Settles.
\newblock Active learning.
\newblock {\em Synthesis Lectures on Artificial Intelligence and Machine
  Learning}, 6(1):1--114, 2012.

\bibitem{thahir2012efficient}
Mohamed Thahir, Tarun Sharma, and Madhavi~K Ganapathiraju.
\newblock An efficient heuristic method for active feature acquisition and its
  application to protein-protein interaction prediction.
\newblock In {\em BMC proceedings}, volume~6, page~S2. BioMed Central, 2012.

\bibitem{tomczak2017vae}
Jakub~M Tomczak and Max Welling.
\newblock Vae with a vampprior.
\newblock {\em arXiv preprint arXiv:1705.07120}, 2017.

\bibitem{valera2017automatic}
Isabel Valera and Zoubin Ghahramani.
\newblock Automatic discovery of the statistical types of variables in a
  dataset.
\newblock In {\em Proceedings of the 34th International Conference on Machine
  Learning-Volume 70}, pages 3521--3529. JMLR. org, 2017.

\bibitem{zhang2018advances}
Cheng Zhang, Judith B{\"u}tepage, Hedvig Kjellstr{\"o}m, and Stephan Mandt.
\newblock Advances in variational inference.
\newblock {\em IEEE transactions on pattern analysis and machine intelligence},
  41(8):2008--2026, 2018.

\end{thebibliography}
\appendix
\newpage
\onecolumn
\section{Additional Derivations}

\subsection{Information reward approximation for hierarchical generative models in the present of missing latent variable}
\label{sec:app_information}

We consider the estimation of the following \emph{information reward} function
\begin{equation*} \label{eq:appendix_IR}
R_I(\mathbf{x}_i,\mathbf{x}_O) = \mathbb{E}_{\mathbf{x}_i \sim p(\mathbf{x}_i|\mathbf{x}_O)}
\mathbb{KL}\left[p(\mathbf{x}_\Phi | \mathbf{x}_i,\mathbf{x}_O) \,\|\, p(\mathbf{x}_\Phi | \mathbf{x}_O)
\right]
\end{equation*}

Using our proposed VAEM method (the partial VAEM version in \ref{sec:partial}). The VAEM is a hierarchical generative model trained by the two-stage procedure described in the paper. Conditional inference of VAEM of missing data follows the following sampling process:
\begin{align*}
     z_d &\sim q_d(z_d|x_{,d},\mathbf{\Phi}_d) \ \ \forall{d} \in O, \ \  \mathbf{z}_O  = \{z_d|d \in O \} \nonumber \\
     \mathbf{h} &\sim q_{\lambda}(\mathbf{h}|\mathbf{z}_O) \nonumber \\
     z_s &\sim p_{\psi}(z_s|\mathbf{h}) \ \ \forall{s} \in U, \ \  \mathbf{z}_U  = \{z_s|s \in U \} \nonumber \\
     x_s &\sim p_{\theta}(x_s|\mathbf{z}_U,\mathbf{z}_O) \ \ \forall{s} \in U, \ \  \mathbf{x}_U  = \{x_s|s \in U \}
 \end{align*}{}

Note that for compactness, we omitted the notation for input $\mathbf{x}_O$ and $\mathbf{x}_i$ to the all partial inference nets $q_\lambda$. Where $\mathbf{z}_O$ is the observed latent variables of marginal VAEs, and $\mathbf{z}_U$ are unobserved. We will use this VAEM to estimate any probabilistic quantities in information reward \ref{eq:appendix_IR}. 

Applying the chain rule of KL-divergence on the term $\mathbb{KL}\left[p(\mathbf{x}_\Phi | \mathbf{x}_i,\mathbf{x}_O) \,\|\, p(\mathbf{x}_\Phi | \mathbf{x}_O)\right]$, we have:
\begin{align*}
&\mathbb{KL}( p(\mathbf{x}_\Phi|\mathbf{x}_i,\mathbf{x}_O)||{p(\mathbf{x}_\Phi|\mathbf{x}_O) })  \\
&= \mathbb{KL}( p(\mathbf{x}_\Phi,\mathbf{z}_i,\mathbf{z}_O,\mathbf{h}|\mathbf{x}_i,\mathbf{x}_O)||p(\mathbf{x}_\Phi, \mathbf{z}_i,\mathbf{z}_O,\mathbf{h}|\mathbf{x}_O)) \\
& - \mathbb{E}_{\mathbf{x}_\Phi \sim p(\mathbf{x}_\Phi|\mathbf{x}_i,\mathbf{x}_O)} \left[ \mathbb{KL}( p(\mathbf{z}_\Phi,\mathbf{z}_i,\mathbf{z}_O,\mathbf{h}|\mathbf{x}_\Phi,\mathbf{x}_i,\mathbf{x}_O)||p(\mathbf{z}_\Phi, \mathbf{z}_i,\mathbf{z}_O,\mathbf{h}|\mathbf{x}_\Phi, \mathbf{x}_O)) \right],
\end{align*}

Based on the independencies of marginal VAEs, we have
$p(\mathbf{x}_\Phi, \mathbf{z}_i,\mathbf{z}_O,\mathbf{h}|\mathbf{x}_O)) = p(\mathbf{x}_\Phi,\mathbf{z}_O,\mathbf{h}|\mathbf{x}_O))p(\mathbf{z}_i)$, $p(\mathbf{z}_\Phi, \mathbf{z}_i,\mathbf{z}_O,\mathbf{h}|\mathbf{x}_\Phi, \mathbf{x}_O)) = p(\mathbf{z}_\Phi, \mathbf{z}_i,\mathbf{z}_O,\mathbf{h}|\mathbf{x}_\Phi, \mathbf{x}_O))p(\mathbf{z}_i)$.

Using again the KL-divergence chain rule on $\mathbb{KL}( p(\mathbf{x}_\Phi,\mathbf{z}_i,\mathbf{z}_O,\mathbf{h}|\mathbf{x}_i,\mathbf{x}_O)||p(\mathbf{x}_\Phi, \mathbf{z}_i,\mathbf{z}_O,\mathbf{h}|\mathbf{x}_O))$, we have:
\begin{align*}
&\mathbb{KL}( p(\mathbf{x}_\Phi,\mathbf{z}_i,\mathbf{z}_O,\mathbf{h}|\mathbf{x}_i,\mathbf{x}_O)||p(\mathbf{x}_\Phi, \mathbf{z}_i,\mathbf{z}_O,\mathbf{h}|\mathbf{x}_O)) \\
&= \mathbb{KL}( p(\mathbf{z}_i,\mathbf{z}_O,\mathbf{h}|\mathbf{x}_i,\mathbf{x}_O)||p( \mathbf{z}_i,\mathbf{z}_O,\mathbf{h}|\mathbf{x}_O))  + \mathbb{E}_{p(\mathbf{z}_\Phi, \mathbf{z}_i,\mathbf{z}_O,\mathbf{h}|\mathbf{x}_{i}, \mathbf{x}_O))} \mathbb{KL}( p(\mathbf{x}_\Phi|\mathbf{z}_i,\mathbf{z}_O,\mathbf{h},\mathbf{x}_i,\mathbf{x}_O)||p(\mathbf{x}_\Phi| \mathbf{z}_i,\mathbf{z}_O,\mathbf{h}, \mathbf{x}_O)) \\ 
&= \mathbb{KL}( p(\mathbf{z}_i,\mathbf{z}_O,\mathbf{h}|\mathbf{x}_i,\mathbf{x}_O)||p( \mathbf{z}_i,\mathbf{z}_O,\mathbf{h}|\mathbf{x}_O))  + \mathbb{E}_{p(\mathbf{z}_\Phi, \mathbf{z}_i,\mathbf{z}_O,\mathbf{h}|\mathbf{x}_{i}, \mathbf{x}_O))} \mathbb{KL}( p(\mathbf{x}_\Phi|\mathbf{z}_i,\mathbf{z}_O,\mathbf{h})||p(\mathbf{x}_\Phi| \mathbf{z}_i,\mathbf{z}_O,\mathbf{h})) \\ 
& = \mathbb{KL}( p(\mathbf{z}_i,\mathbf{z}_O,\mathbf{h}|\mathbf{x}_i,\mathbf{x}_O)||p(\mathbf{z}_i,\mathbf{z}_O,\mathbf{h}|\mathbf{x}_O)).
\end{align*}
Note that the last two equalities does not hold for the discriminative version of VAEM described in Section \ref{sec:predictive}. Fortunately, $\mathbb{E}_{\mathbf{x}_i \sim p(\mathbf{x}_i|\mathbf{x}_O)} \mathbb{KL}( p(\mathbf{x}_\Phi|\mathbf{z}_i,\mathbf{z}_O,\mathbf{h},\mathbf{x}_i,\mathbf{x}_O)||p(\mathbf{x}_\Phi| \mathbf{z}_i,\mathbf{z}_O,\mathbf{h}, \mathbf{x}_O)) = 0$ still holds for the discriminative version, hence we will still arrive at the same result.

The KL-divergence term in the reward formula is now rewritten as follows,
\begin{align*}
&\mathbb{KL}( p(\mathbf{x}_\Phi|\mathbf{x}_i,\mathbf{x}_O)||p(\mathbf{x}_\Phi|\mathbf{x}_O))  \\
& =  {\color{blue}\mathbb{KL}( p(\mathbf{z}_i,\mathbf{z}_O,\mathbf{h}|\mathbf{x}_i,\mathbf{x}_O)||p(\mathbf{z}_i,\mathbf{z}_O,\mathbf{h}|\mathbf{x}_O))} \\
& - {\color{red}\mathbb{E}_{\mathbf{x}_\Phi \sim p(\mathbf{x}_\Phi|\mathbf{x}_i,\mathbf{x}_O)} \left[ \mathbb{KL}( p(\mathbf{z}_{\Phi},\mathbf{z}_i,\mathbf{z}_O,\mathbf{h}|\mathbf{x}_\Phi,\mathbf{x}_i,\mathbf{x}_O)||p(\mathbf{z}_{\Phi}, \mathbf{z}_i,\mathbf{z}_O,\mathbf{h}|\mathbf{x}_\Phi, \mathbf{x}_O)) \right]}.
\end{align*}

For the term in blue, we have:
\begin{align*}
     &{\color{blue}\mathbb{KL}( p(\mathbf{z}_i,\mathbf{z}_O,\mathbf{h}|\mathbf{x}_i,\mathbf{x}_O)||p(\mathbf{z}_i,\mathbf{z}_O,\mathbf{h}|\mathbf{x}_O))} \\
     &=\mathbb{KL}( p(\mathbf{z}_i,\mathbf{z}_O|\mathbf{x}_i,\mathbf{x}_O)||p(\mathbf{z}_O|\mathbf{x}_O)p(\mathbf{z}_i)) \\
     & + \mathbb{E}_{\mathbf{z}_i,\mathbf{z}_O \sim p(\mathbf{z}_i,\mathbf{z}_O|\mathbf{x}_i,\mathbf{x}_O)} \left[ \mathbb{KL}\left( p(\mathbf{h}|\mathbf{z}_i,\mathbf{z}_O)||p(\mathbf{h}|\mathbf{z}_O) \frac{p(\mathbf{z}_i)}{p(\mathbf{z}_i|\mathbf{x}_O)} \right) \right] \\
     &=\mathbb{KL}( p(\mathbf{z}_i|\mathbf{x}_i)||p(\mathbf{z}_i)) + \mathbb{E}_{\mathbf{z}_i,\mathbf{z}_O \sim p(\mathbf{z}_i,\mathbf{z}_O|\mathbf{x}_i,\mathbf{x}_O)} \left[ \mathbb{KL}\left( p(\mathbf{h}|\mathbf{z}_i,\mathbf{z}_O)||p(\mathbf{h}|\mathbf{z}_O) \right) \right]
\end{align*}{}

Similarly for the term in red, we have:
\begin{align*}
     &{\color{red} \mathbb{KL}( p(\mathbf{z}_{\Phi},\mathbf{z}_i,\mathbf{z}_O,\mathbf{h}|\mathbf{x}_\Phi,\mathbf{x}_i,\mathbf{x}_O)||p(\mathbf{z}_{\Phi},\mathbf{z}_i,\mathbf{z}_O,\mathbf{h}|\mathbf{x}_\Phi,\mathbf{x}_O))} \\
     &=\mathbb{KL}( p(\mathbf{z}_{\Phi},\mathbf{z}_i,\mathbf{z}_O|\mathbf{x}_\Phi,\mathbf{x}_i,\mathbf{x}_O)||p(\mathbf{z}_{\Phi},\mathbf{z}_O|\mathbf{x}_\Phi,\mathbf{x}_O)p(\mathbf{z}_i)) \\
     & + \mathbb{E}_{\mathbf{z}_{\Phi},\mathbf{z}_i,\mathbf{z}_O \sim p(\mathbf{z}_{\Phi},\mathbf{z}_i,\mathbf{z}_O|\mathbf{x}_\Phi,\mathbf{x}_i,\mathbf{x}_O)} \left[ \mathbb{KL}\left( p(\mathbf{h}|\mathbf{z}_{\Phi},\mathbf{z}_i,\mathbf{z}_O)||p(\mathbf{h}|\mathbf{z}_{\Phi},\mathbf{z}_O) \frac{p(\mathbf{z}_i)}{p(\mathbf{z}_i|\mathbf{x}_\Phi,\mathbf{x}_O)} \right) \right] \\
     &=\mathbb{KL}(\mathbf{z}_i|\mathbf{x}_i)||p(\mathbf{z}_i)) + \mathbb{E}_{\mathbf{z}_{\Phi},\mathbf{z}_i,\mathbf{z}_O \sim p(\mathbf{z}_{\Phi},\mathbf{z}_i,\mathbf{z}_O|\mathbf{x}_\Phi,\mathbf{x}_i,\mathbf{x}_O)} \left[ \mathbb{KL}\left( p(\mathbf{h}|\mathbf{z}_{\Phi},\mathbf{z}_i,\mathbf{z}_O)||p(\mathbf{h}|\mathbf{z}_{\Phi},\mathbf{z}_O) \right) \right]
\end{align*}{}

Finally, we have:
\begin{align*} 
&R_I(\mathbf{x}_i,\mathbf{x}_O) \nonumber \\
=& \mathbb{E}_{\mathbf{x}_i \sim p(\mathbf{x}_i|\mathbf{x}_O)}
\mathbb{KL}\left[p(\mathbf{x}_\Phi | \mathbf{x}_i,\mathbf{x}_O) \,\|\, p(\mathbf{x}_\Phi | \mathbf{x}_O)
\right] \\
=&\mathbb{E}_{\mathbf{x}_i \sim p(\mathbf{x}_i|\mathbf{x}_O)} {\color{blue}\mathbb{KL}( p(\mathbf{z}_i,\mathbf{z}_O,\mathbf{h}|\mathbf{x}_i,\mathbf{x}_O)||p(\mathbf{z}_i,\mathbf{z}_O,\mathbf{h}|\mathbf{x}_O))} \\
- & \mathbb{E}_{\mathbf{x}_i \sim p(\mathbf{x}_i|\mathbf{x}_O)} {\color{red}\mathbb{E}_{\mathbf{x}_\Phi \sim p(\mathbf{x}_\Phi|\mathbf{x}_i,\mathbf{x}_O)} \left[ \mathbb{KL}( p(\mathbf{z}_{\Phi},\mathbf{z}_i,\mathbf{z}_O,\mathbf{h}|\mathbf{x}_\Phi,\mathbf{x}_i,\mathbf{x}_O)||p(\mathbf{z}_{\Phi}, \mathbf{z}_i,\mathbf{z}_O,\mathbf{h}|\mathbf{x}_\Phi, \mathbf{x}_O)) \right]} \\
= & \mathbb{E} _{ \mathbf{x}_i,\mathbf{z}_i,\mathbf{z}_O\sim p(\mathbf{x}_i,\mathbf{z}_i,\mathbf{z}_O|\mathbf{x}_O)} \left\{ \mathbb{KL}\left[ p(\mathbf{h}|\mathbf{z}_i,\mathbf{z}_O)||p( \mathbf{h}|\mathbf{z}_O)\right]\right. \\ \nonumber
-&  \left. \mathbb{E} _{\mathbf{x}_\Phi,\mathbf{z}_\Phi\sim p(\mathbf{x}_\Phi,\mathbf{z}_\Phi, |\mathbf{x}_O)}  \mathbb{KL}\left[ p(\mathbf{h}|\mathbf{z}_{\Phi},\mathbf{z}_i,\mathbf{z}_O)||p( \mathbf{h}|\mathbf{z}_{\Phi}, \mathbf{z}_O) \right] \right\}.
\end{align*}

We can then plug in the VAEM model distirbutions: 
\begin{align*}
&p(\mathbf{x}_i,\mathbf{z}_i,\mathbf{z}_O|\mathbf{x}_O) = p_{\theta,\phi}(\mathbf{x}_i,\mathbf{z}_i,\mathbf{z}_O|\mathbf{x}_O) \\
& p(\mathbf{x}_\Phi,\mathbf{z}_\Phi, |\mathbf{x}_O) = p_{\theta,\phi}(\mathbf{x}_\Phi,\mathbf{z}_\Phi, |\mathbf{x}_O) \\
& p(\mathbf{h}|\mathbf{z}_i,\mathbf{z}_O) \approx q_\lambda(\mathbf{h}|\mathbf{z}_i,\mathbf{z}_O)\\
& p( \mathbf{h}|\mathbf{z}_O) \approx q_\lambda( \mathbf{h}|\mathbf{z}_O)\\
& p(\mathbf{h}|\mathbf{z}_{\Phi},\mathbf{z}_i,\mathbf{z}_O) \approx q_\lambda(\mathbf{h}|\mathbf{z}_{\Phi},\mathbf{z}_i,\mathbf{z}_O)\\
& p( \mathbf{h}|\mathbf{z}_{\Phi}, \mathbf{z}_O) \approx q_\lambda( \mathbf{h}|\mathbf{z}_{\Phi}, \mathbf{z}_O)\\
\end{align*}
Finally, the information reward is now approximated as:
\begin{align*} 
&R_I(\mathbf{x}_i,\mathbf{x}_O) \nonumber \\
\approx & \mathbb{E} _{ \mathbf{x}_i,\mathbf{z}_i,\mathbf{z}_O\sim p_{\theta,\phi}(\mathbf{x}_i,\mathbf{z}_i,\mathbf{z}_O|\mathbf{x}_O)} \left\{ \mathbb{KL}\left[ q_\lambda(\mathbf{h}|\mathbf{z}_i,\mathbf{z}_O)||q_\lambda( \mathbf{h}|\mathbf{z}_O)\right]\right. \\ \nonumber
-&  \left. \mathbb{E} _{\mathbf{x}_\Phi,\mathbf{z}_\Phi\sim p_{\theta,\phi}(\mathbf{x}_\Phi,\mathbf{z}_\Phi, |\mathbf{x}_O)}  \mathbb{KL}\left[ q_\lambda(\mathbf{h}|\mathbf{z}_{\Phi},\mathbf{z}_i,\mathbf{z}_O)||q_\lambda( \mathbf{h}|\mathbf{z}_\Phi, \mathbf{z}_O) \right] \right\}.
\end{align*}

\subsection{VAEM optimizes a lower bound of joint model log-likelihood} \label{app:proof_vaem}

Next, we show that VAEM improves a valid lower bound of the true log-likelihood. Recall that in the first stage, $D$ individual VAEs are trained independently, i.e. each one is trained to fit a single dimension $x_{nd}$ from the dataset:
\begin{align}
     & (\theta_d^\star,\phi_d^\star) =  {\arg \max}_{\theta_d,\phi_d} \nonumber\\
     & \sum_n \mathbb{E}_{q_{\phi_d}(z_{nd}|x_{nd})} \log\frac{ p_{\theta_d}(x_{nd},z_{nd})}{q_{\phi_d}(z_{nd}|x_{nd})}
\ &\forall{d} \in \{1,2,...,D\},\label{eq:proof_stage_one}
\end{align}

Together, the $D$ individual VAEs define a joint distribution over $\mathbf{x}_n$: $\log p_{\theta}(\mathbf{x}_{n}) := \log \sum_{\mathbf{z}_n} \prod_d p_{\theta_d}(\mathbf{x}_{n}|\mathbf{z}_{n}) p_0(\mathbf{z}_{n})$, where $p_0(\mathbf{z}_{n})$ is a factorized standard normal distribution. Note that $\mathbb{E}_{q_{\phi_d}(z_{nd}|x_{nd})} \log\frac{ p_{\theta_d}(x_{nd},z_{nd})}{q_{\phi_d}(z_{nd}|x_{nd})}$ in Equation \ref{eq:proof_stage_one} is a lower bound of $\log p_{\theta_d}(\mathbf{x}_{nd})$, therefore stage one jointly optimizes a valid lower bound of $\log p_{\theta}(\mathbf{x}_{n})$:
\begin{align}
    \log p_{\theta}(\mathbf{x}_{n}) \geq \sum_d \mathbb{E}_{q_{\phi_d}(z_{nd}|x_{nd})} \log\frac{ p_{\theta_d}(x_{nd},z_{nd})}{q_{\phi_d}(z_{nd}|x_{nd})} \label{eq:first_elbo}
\end{align}
where $p_{\theta_d}(x_{nd},z_{nd}) = p_{\theta_d}(x_{nd}|z_{nd}) p_0(z_{nd})$.

Then we proceed to the second stage. In this stage, the dependency VAE $p_{\psi}(\mathbf{z}) = \mathbb{E}_{p(\mathbf{h})} p_{\psi}(\mathbf{z}|\mathbf{h})$, is trained on the latent representations $\mathbf{z}$ provided by the encoders of the marginal VAEs in the first stage:
 \begin{align}
     \mathbf{x}_{n} &\sim p_{\text{data}}(\mathbf{x}) ,\nonumber \\
     z_{nd} &\sim q_{\mathbf{\phi}_d}(z_d|x_{nd}), \ \ \forall{d} \in \{1,...,D\},\nonumber \\
      (\psi^\star,\lambda^\star) &\propto  \arg \max_{(\psi,\lambda)} \sum_n \mathbb{E}_{q_{\lambda}(\mathbf{h}_n|\mathbf{z}_n,\mathbf{x}_{n})} \log\frac{ p_{\psi}(\mathbf{z}_n,\mathbf{h}_n)}{q_\lambda(\mathbf{h}_n|\mathbf{z}_n,\mathbf{x}_{n})}.
      \label{eq:proof_stage_two}
 \end{align}
 
 In other words, the second stage of VAEM improves $p_0(\mathbf{z})$ (a factorized standard Gaussian) by $p_\psi(\mathbf{z})$. Since we optimizes the ELBO of $p_\psi(\mathbf{z})$, if we can show that 
 \begin{align}
     \mathbb{E}_{q_{\lambda^\star}(\mathbf{h}|\mathbf{z},\mathbf{x})} \log\frac{ p_{\psi^\star}(\mathbf{z},\mathbf{h})}{q_{\lambda^\star}(\mathbf{h}|\mathbf{z},\mathbf{x})} >\log p_0(\mathbf{z}) \label{eq:assumption}
 \end{align}

 Then we can conclude that the second stage will improve the lower bound given by the first stage (Equation \ref{eq:first_elbo}). Next, we show that Equation \ref{eq:assumption} indeed holds. All we need to do is to initialize $p_{\psi_0}(\mathbf{z})$ so that $p_{\psi_0}(\mathbf{z}) = p_0(\mathbf{z})$, and initialize $q_{\lambda_0}(\mathbf{h}|\mathbf{z},\mathbf{x})$ so that $q_{\lambda_0}(\mathbf{h}|\mathbf{z},\mathbf{x})$ is the exact posterior of $p_{\psi_0}(\mathbf{h}|\mathbf{z})$. 
 
 Note that it is trivial to show that such initialization is possible. One way to do this is to use relu activation functions for hidden layers in the dependency VAE, and then initialize all the weights biases, log variances in the decoder and encoders to be zero. In this way, the decoder of dependency VAE will ignores the latent variable $\mathbf{h}$ and generates factorized standard Gaussians. The encoder with zero initialization will also give factorized Gaussian, which will be identical to the prior $p(\mathbf{h})$. This is exactly the true posterior $p_{\psi_0}(\mathbf{h}|\mathbf{z})$, since the dependency network decoder completely ignores its input: $p_{\psi_0}(\mathbf{z}|\mathbf{h}) = p_{\psi_0}(\mathbf{z})$.  
 
 Note that there are many ways to achieve the Equation \ref{eq:assumption}, the above is only one way to do this. The above zero initialization setting is exactly what we have used in our experiments. Finally, we can ensure that by optimizing Equation \ref{eq:proof_stage_two}, we always have:
 \begin{align}
     & \sum_n \log p_{\psi^\star}(\mathbf{z}_n) \geq \sum_n  \mathbb{E}_{q_{\lambda^\star}(\mathbf{h}_n|\mathbf{z}_n,\mathbf{x}_n)} \log\frac{ p_{\psi^\star}(\mathbf{z}_n,\mathbf{h}_n)}{q_{\lambda^\star}(\mathbf{h}_n|\mathbf{z}_n,\mathbf{x}_n)} \label{eq:init} \\
     & > \sum_n \mathbb{E}_{q_{\lambda_0}(\mathbf{h}_n|\mathbf{z}_n,\mathbf{x}_{n})} \log\frac{ p_{\psi_0}(\mathbf{z}_n,\mathbf{h}_n)}{q_{\lambda_0}(\mathbf{h}_n|\mathbf{z}_n,\mathbf{x}_{n})} = \sum_n \log p_{\psi_0}(\mathbf{z}_n) = \sum_n \log p_0(\mathbf{z}_n) \nonumber
 \end{align}
 
 Therefore, we finally have:
 
 \begin{align}
    &  \sum_n  \log \sum_{\mathbf{z_n}} \prod_d  p_{\theta_d}(x_{nd}|z_{nd})  p_{\psi^\star}(\mathbf{z}_n)  \nonumber \\
    & \geq \sum_n \mathbb{E}_{\prod_dq_{\phi_d}(z_{nd}|x_{nd})} \sum_d  \log\frac{ p_{\theta_d}(x_{nd}|z_{nd})}{q_{\phi_d}(z_{nd}|x_{nd})} + \sum_n  \mathbb{E}_{\prod_d q_{\phi_d}(z_{nd}|x_{nd})} \mathbb{E}_{q_{\lambda^\star}(\mathbf{h}_n|\mathbf{z}_n,\mathbf{x}_n)} \log\frac{ p_{\psi^\star}(\mathbf{z}_n,\mathbf{h}_n)}{q_{\lambda^\star}(\mathbf{h}_n|\mathbf{z}_n,\mathbf{x}_n)} \nonumber \\
    & >  \sum_n  \sum_d \mathbb{E}_{q_{\phi_d}(z_{nd}|x_{nd})}  \log\frac{ p_{\theta_d}(x_{nd},z_{nd})}{q_{\phi_d}(z_{nd}|x_{nd})}
 \end{align}

Where the second row is the ELBO after the second stage, and the third row is the ELBO after the first stage. The first inequality follows from Jensen's inequality, and the second inequality follows from Equation \ref{eq:init}. Therefore, the two stage procedure of VAEM always increases the lower bound of true log-likelihood.

\qed

\paragraph{Two stage training vs joint training}
To summarize, the VAEM training procedure optimizes the following ELBO in a two stage manner:
\begin{align}
 &  \sum_n  \log \sum_{\mathbf{z_n}} \prod_d  p_{\theta_d}(x_{nd}|z_{nd})  p_{\psi}(\mathbf{z}_n) \geq \nonumber \\
    &
  \sum_n \mathbb{E}_{\prod_dq_{\phi_d}(z_{nd}|x_{nd})} \sum_d  \log\frac{ p_{\theta_d}(x_{nd}|z_{nd})}{q_{\phi_d}(z_{nd}|x_{nd})} + \sum_n  \mathbb{E}_{\prod_d q_{\phi_d}(z_{nd}|x_{nd})} \mathbb{E}_{q_{\lambda}(\mathbf{h}_n|\mathbf{z}_n,\mathbf{x}_n)} \log\frac{ p_{\psi}(\mathbf{z}_n,\mathbf{h}_n)}{q_{\lambda}(\mathbf{h}_n|\mathbf{z}_n,\mathbf{x}_n)}  \label{eq:vaem_final_objective}
\end{align}

In the first stage, it optimizes Equation \ref{eq:vaem_final_objective}, but initialize $p_{\psi}(\mathbf{z}_n)$ to standard gaussian $p_0(\mathbf{z}_n)$ and keep it fixed. In the second stage, VAEM also optimizes Equation \ref{eq:vaem_final_objective}, but now keeps $p_{\theta_d}(x_{nd}|z_{nd})$ and $q_{\phi_d}(z_{nd}|x_{nd})$ fixed, and optimizes $p_{\psi}(\mathbf{z}_n,\mathbf{h}_n)$ and $q_{\lambda}(\mathbf{h}_n|\mathbf{z}_n,\mathbf{x}_n)$.  

Note that if we directly optimize Equation \ref{eq:vaem_final_objective} jointly instead of the two stage method, then $q_{\phi_d}(z_{nd}|x_{nd})$ will not be regularized by standard Gaussian prior $p_0(\mathbf{z}_{nd})$. As a result, we will lose the uniformity of the $z_{nd}$ ( to exact, the uniformity of aggregated posterior $\frac{1}{N}\sum_n q_{\phi_d}(z_{nd}|x_{nd})$).

On the contrary, in our two stage method, the latent representations $z_d$ provided by the marginal VAEs will have uniform
properties across dimensions. Each of these variables is encouraged to be close to a standard normal distribution thanks to the regularization effect from the Gaussian prior $p_0(z_{nd})$. In this way, we sidestep the heterogeneous mixed-type problem and the dependency VAE can just focus on learning dependencies among single-type homogeneous variables.

\section{Enhancing predictive performance of VAEM: training procedure} \label{sec:app_predictive}
In order to enhance the predictive performance of VAEM, the following alternative factorization is proposed:
\begin{equation*}
    p_\theta(\mathbf{x}_O, \mathbf{x}_\Phi) =
\mathbb{E}_{\mathbf{x}_{U\setminus \Phi},\mathbf{h} \sim p_\theta(\mathbf{x}_{U\setminus \Phi},\mathbf{h}|\mathbf{x}_O)} p_\gamma(\mathbf{x}_\Phi|\mathbf{x}_O, \mathbf{x}_{U\setminus \Phi}, \mathbf{h})p_\theta(\mathbf{x}_O)
\end{equation*}{}

For compactness, the notation for input $\mathbf{x}_O$ and $\mathbf{x}_i$ to the all partial inference nets $q_\lambda$ will be omitted. Note that, to train this model, we also need data samples of $\mathbf{x}_\Phi$ during training (however $\mathbf{x}_\Phi$ will not be observed during active learning task). This model is trained using the following procedure:
\begin{itemize}
    \item Train a partial VAEM on $\mathbf{x}_O$ ($\mathbf{x}_\Phi \cap \mathbf{x}_O = \emptyset$) using the two-stage method described in Section \ref{sec:MHM}. Now we have a graphical model induced by the model $p_\theta(\mathbf{x}_O)$.
    \item Expand the graph by adding the node $\mathbf{x}_\Phi$ to the graph. Now the joint distribution is defined as $p_\theta(\mathbf{x}_O, \mathbf{x}_\Phi) =
\mathbb{E}_{\mathbf{x}_{U\setminus \Phi},\mathbf{h} \sim p_\theta(\mathbf{x}_{U\setminus \Phi},\mathbf{h}|\mathbf{x}_O)} p_\gamma(\mathbf{x}_\Phi|\mathbf{x}_O, \mathbf{x}_{U\setminus \Phi}, \mathbf{h}) p_\theta(\mathbf{x}_O)$. Note that no new parameters need to be introduced for the partial inference net of the dependency network $q_\lambda(\mathbf{h}|\mathbf{z}_O,\mathbf{z}_\Phi)$, since the partial inference net automatically handles inputs with different dimensionalities. 

\item Define the marginal VAE encoder for $x_\Phi$ as $q_d(z_{\Phi}|x_{n,\Phi},\mathbf{\phi}_\Phi) = \delta(z_{\Phi}-x_{\Phi})$, and the decoder to be $p_d(x_{n,\Phi}|z_d,\mathbf{\theta}_\Phi) = \delta(x_{\Phi}-z_{\Phi})$ (i.e., both are identity deterministic mappings).
\item The partial inference net parameters of the dependency network can be updated by the following procedure:
 \begin{align*}
     z_d &\sim q_d(z_d|x_{{\text{data}},d},\mathbf{\phi}_d) \ \ \forall{d} \in O\cup \Phi, \ \  \mathbf{z}_{O\cup \Phi}  = \{z_d|d \in O\cup \Phi \} \nonumber \\
     \Delta \lambda &\propto  \nabla_{\lambda} \mathbb{E}_{q_{\lambda}(\mathbf{h}|\mathbf{z}_{O\cup \Phi})} \left[\log\frac{ \prod_{d\in O}p_{\psi}(z_d|\mathbf{h})p(\mathbf{h})}{q_\lambda(\mathbf{h}|\mathbf{z}_{O\cup \Phi})} + \mathbb{E}_{\mathbf{x}_{U\setminus \Phi} \sim p_{\theta,\psi}(\mathbf{x}_{U\setminus \Phi}|\mathbf{h})} \log p_\gamma(\mathbf{x}_\Phi|\mathbf{x}_O, \mathbf{x}_{U\setminus \Phi}, \mathbf{h})\right]
 \end{align*}{}

\item The the parameters for $p_\gamma(\mathbf{x}_\Phi|\mathbf{x}_O, \mathbf{x}_{U\setminus \Phi}, \mathbf{h})$ can be updated by the following procedure:
\begin{align*}
     z_d &\sim q_d(z_d|x_{,d},\mathbf{\phi}_d) \ \ \forall{d} \in O, \ \  \mathbf{z}_O  = \{z_d|d \in O \} \nonumber \\
     \mathbf{h} &\sim q_{\lambda}(\mathbf{h}|\mathbf{z}_O) \nonumber \\
     z_s &\sim p_{\psi}(z_s|\mathbf{h}) \ \ \forall{s} \in U\setminus \Phi, \ \  \mathbf{z}_{U\setminus \Phi }  = \{z_s|s \in U\setminus \Phi \} \nonumber \\
     x_s &\sim p_{\theta}(x_s|\mathbf{z}_U,\mathbf{z}_O) \ \ \forall{s} \in U\setminus \Phi, \ \  \mathbf{x}_{U \setminus \Phi }  = \{x_s|s \in U\setminus \Phi \} \\
     \gamma^\star &= \arg \max _\gamma \log p_\gamma(\mathbf{x}_\Phi|\mathbf{x}_O, \mathbf{x}_{U\setminus \Phi}, \mathbf{h})
 \end{align*}{}
\end{itemize}{}

\section{Additional Experiment Settings} \label{app:exp_set}

subsection{Datasets details}
We use the same collection of mixed type datasets in all tasks:
\begin{compactitem}
    \item  Two standard UCI benchmark datasets: Boston housing (13 continuous, 1 categorical) and energy efficiency (6 continuous, 3 categorical) \cite{Dua:2017};
    \item Two relatively large real-world dataset: Bank marketing (45211 instances, 11 continuous, 8 categorical, 2 discrete); \cite{moro2011using} and Avocado sales prediction (18249 instances, 9 continuous, 5 categorical).
    \item One real-world medical dataset: Medical Information Mart for Intensive Care (MIMIC III) \cite{johnson2016mimic}, the largest public medical dataset containing records of 21139 patients (after processing following \cite{harutyunyan2017multitask}). We focus on the mortality prediction task based on 17 medical instruments (13 continuous, 4 categorical). Since the dataset is imbalanced (over 80 \% of the data has mortality $= 0$), we balance the dataset by down-sampling the majority class. The time-series observations are averaged to obtain iid data points.
\end{compactitem}{}

\subsection{Additional model specification} \label{app:exp_set_model}

\subsubsection{Baselines: general information}
We have used the following baselines in our experiments:
\begin{compactitem}
    \item Heterogeneous-Incomplete VAE (HI-VAE) \cite{nazabal2018handling}. We adopt the multi-head structure of HI-VAE and match the dimensionality of latent variables to be the same as our \name. HI-VAE is an important baseline, since it is motivated in a similar way as our VAEM, but all marginal VAEs are trained jointly rather as opposed to our two-stage method. We denote this by \texttt{VAE-HI}
    
    \item VAE: A vanilla VAE equipped with a VampPrior \cite{tomczak2017vae}. The number of latent dimensions is the same as in the second stage of VAEM. We denote this by \texttt{VAE}. 
    
    \item VAE with extended latent dimension: Note that the total number of latent variables of \name is $D+L$, where $D$ and $L$ are the dimensionalities of the data and the latent space, respectively. This baseline is like the previous one, but with the latent dimension given by $D+L$. We denote this baseline by \texttt{VAE-extended}.
    
    \item VAE with automatically balanced likelihoods. This baseline tries to automatically equal the scale of each likelihood term of the different variable types in the ELBO by multiplying each likelihood term with an {adaptive constant (Appendix \ref{app:exp_set_model})}. We denote this baseline by \texttt{VAE-balanced}. 
\end{compactitem}{}

\subsubsection{Baseline: VAE with balanced likelihoods} This baseline is a naive strategy that tries to automatically balance the scale of the log-likelihood values of different variable types in the ELBO, by adaptively multiplying a constant before likelihood terms. More specifically, consider the variational lower bound (ELBO) of vanilla VAE:
\begin{equation*}
\begin{aligned}
& \log p_\theta(\mathbf{x})  \geq \mathbb{E}_{q_\phi(\mathbf{z}|\mathbf{x})} \log\frac{ p_\theta(\mathbf{x},\mathbf{z})}{q_\phi(\mathbf{z}|\mathbf{x})} \\ 
& = \sum_{s\in \mathcal{P}}\mathbb{E}_{q_\phi(\mathbf{z}|\mathbf{x})} \log\frac{ p_\theta(\mathbf{x}_{s\in \mathcal{P}},\mathbf{z})}{q_\phi(\mathbf{z}|\mathbf{x})}
\end{aligned}
\end{equation*}
Where $\mathcal{P}$ is the set of variable types (e.g., continuous, categorical), and $\mathbf{x}_s$ is the set of variables that belong to $s$-th type. In VAE with balanced likelihoods, we weight each likelihood terms by $\{\beta_1,\beta_2,...,\beta_|\mathbf{P}|\}$:
\begin{equation*}
\begin{aligned}
& \sum_{s\in \mathcal{P}} \beta_s \mathbb{E}_{q_\phi(\mathbf{z}|\mathbf{x})} \log\frac{ p_\theta(\mathbf{x}_{s\in \mathcal{P}},\mathbf{z})}{q_\phi(\mathbf{z}|\mathbf{x})}
\end{aligned}
\end{equation*}

Where $\sum_s \beta_s = 1$, such that:
\begin{equation*}
\beta_s \mathbb{E}_{q_\phi(\mathbf{z}|\mathbf{x})} \log p_\theta(\mathbf{x}_{s}|\mathbf{z}) = \beta_t \mathbb{E}_{q_\phi(\mathbf{z}|\mathbf{x})} \log p_\theta(\mathbf{x}_{t}|\mathbf{z}), \ \  \forall{s,t} \in \mathcal{P}
\end{equation*}

In practice, at each epoch of training, a mini-batch $\{\mathbf{x}_j\}_{1\leq j \leq J}$ is selected, and $\beta_s$ are estimated such that:
\begin{equation*}
\beta_s \sum_j \mathbb{E}_{q_\phi(\mathbf{z}_j|\mathbf{x}_j)} \log p_\theta(\mathbf{x}_{j,s}|\mathbf{z}_j) = \beta_t \sum_j \mathbb{E}_{q_\phi(\mathbf{z}_j|\mathbf{x}_j)} \log p_\theta(\mathbf{x}_{j,t}|\mathbf{z}_j), \ \  \forall{s,t} \in \mathcal{P}
\end{equation*}

\subsubsection{Likelihood function specification}
In this paper, we consider three variable types: continuous, categorical, and discrete. For continuous and categorical variables, we follow the specification of \cite{nazabal2018handling}. In other words, to specify the likelihood function of all VAE decoders $p_{\theta_d}(x_{nd}|z_{nd})$ in our paper, we use Gaussian likelihood with constant observational noises $p_{\theta_d}(x_{nd}|z_{nd}) = \mathcal{N}(x_{nd};\mu(z_{nd}),\sigma^2)$ for continuous data; and for categorical data, we use categorical likelihood with one-hot representation $p_{\theta_d}(x_{nd}|z_{nd}) = \langle \mathbf{l}(z_{nd}),\texttt{one-hot}(x_{nd}) \rangle$, where $\mathbf{l}(z_{nd})$ is soft-max output of the decoder.

For discrete variables, we consider two different scenarios: continuous-discrete and ordinal-discrete. Continuous-discrete means that the variable is continuous by its nature, but only discretized values are recorded. For example, the salary (dollars) is a continuous variable, but in practice only discretized values (5000 dollars, 6000 dollars, etc.) are recorded. For this type of variables, we still use Gaussian likelihood, but the decoder output will be rounded to the closest discrete value. On the other hand, ordinal-discrete variables (such as ratings) are the ones with natural orderings, and the distance between each value is not known. For ordinal variables, we use ordinal regression likelihood used in \cite{paquet2012hierarchical}.

Note that the above settings are used for all models including VAEM and other baselines.

\subsubsection{Partial inference net with non-continuous input}
In section \ref{sec:partial}, the partial inference net $q_\lambda(\mathbf{h}|\mathbf{z}_O,\mathbf{x}_O)$ is constructed based on the element-wise multiplication operation $\mathbf{s}_O := \{ v \times \mathbf{e}_v|v \in \mathbf{z}_O \cup \mathbf{x}_O\}$. How is $v\times \mathbf{e}_v$ defined if $v$ is non-continuous? For categorical and ordinal-discrete variable for example, the operation $v\times\mathbf{e}_v$ is defined as 
\begin{equation*}
    v\times\mathbf{e}_v: = vec( \texttt{one-hot}(v)\otimes \mathbf{e}_v)
\end{equation*}{}
Where $\otimes$ is outer-product between vectors, \texttt{one-hot} is the one-hot representation of the categorical/ordinal variables, and $vec(\cdot)$ is the vectorization operation of a matrix (in other words, it flattens the matrix into a vector).

\subsection{Network structure and hyper parameter settings}\label{app:exp_set_hyper}

\paragraph{Network structures}
All models (except for the marginal VAEs of VAEM and the decoder of HI-VAE) share the same network structures with 20 dimensional diagonal Gaussian latent variables: the generator (decoder) is a 20-50-100 fully connected neural network with ReLU activation functions on hidden units (where $D$ is the data dimension). Note that we use sigmoid activation function for output layer, to reflect our data preprocessing (all data are normalized to between 0 and 1). One exception is the output layer of dependency network of VAEM, where we did not add any activation functions since the scales of the latent variables $z_d$ from marginal VAEs are unknown. The encoders share the same structure of $D$-500-200-40 that maps the observed data into distributional parameters of the latent space. Additionally, we use a $K=$100 dimensional feature mapping parameterized by a single layer neural network, and $M=$10 dimensional feature embedding for each variable. We choose the permutation invariant operator $g$ to be the summation operator. The discriminator described in section \ref{sec:predictive} is a neural network with two layers, each of which has 100 hidden units.

For marginal VAEs of our VAEM, we use 1-dimensional latent variable for each variable.The decoder of marginal VAEs is a 1-50-V single layer neural network, and the encoder network structure is V-50-2, where $V$ is the dimension of the corresponding variable, which is defined to be 1 if the variable is continuous. Otherwise, $V$ is the dimension of the one-hot representation. The same structure is used for the multi-head decoder structure for HI-VAE baseline.

\paragraph{Hyperparameters}  To train our models, we apply Adam optimization \cite{adam2015} with learning rate of 0.001 and a batch size of 100. When the training set is fully observed, We manually generate partially observed version of it by adding artificially missingness at random in the training dataset during training. This will help the model to learn to generate conditional data given observations. We first draw a missing rate parameter from a uniform distribution $\mathcal{U}(0, 1)$ and randomly choose variables as unobserved. This step is repeated at each iteration. We train our models for 3000 full epochs, except for Bank dataset where we used 5000 epochs. For continuous variables, the constant observational noise variance level for Gaussian likelihood functions of decoders are set to be 0.02 (except for MIMIC dataset where we have used 0.3). During evaluation, we use importance sampling with 10K samples to estimate the log-likelihoods for conditional data generation.

\paragraph{Sequential active information acquisition} For SAIA tasks, we use 10 monte-carlo samples from VAE models to estimate reward functions. Since the focus of this paper is comparing the performance of different generative models on heterogeneous mixed type data, we use the SING strategy \cite{ma2018eddi} for SAIA, which uses the objective function as in Equation (\ref{eq:reward_approx}) to find the optimal ordering, by averaging over all the currently observed test data.

\subsection{Additional experiment pipeline setup}
In Section \ref{sec:generation}, during training, the range of all variables is scaled to be between 0 and 1. This transformation is removed when making predictions on the target variables. 

In Section \ref{sec:exp_conditional}, to train these partial models on data with missing
values, we randomly sample 90\% of the dataset to be the training set, and assume that a random fraction (uniformly sampled between 0\% and 99\%) of feature values are missing on each epoch during training. Then, during test time, we assume that 50\% of the test set is observed, and use generative models to infer the unobserved data.

\newpage
\section{Additional experimental results}\label{app:exp_results}

\subsection{Imputation errors of conditional data generation experiment}

Here we also provide results that uses imputation errors to evaluate model performance in Section \ref{sec:exp_conditional}.  

Note that one issue with imputation error is: since now we have mixed type data, the errors of different variables are not directly comparable. Therefore, one often need to introduce a coefficient to weight the error of different types of variables. The ranking of imputation performance will be highly dependent on the choice of such coefficient. 

Here, we set the weighting coefficients to be 1, and calculate the imputation error based on RMSE. For continuous variables, the RMSE/SE is defined as usual; for categorical variables, the RMSE/SE will be calculated based on their one-hot encodings. Then, we take the average of errors across all variables as our final metric. The calculation are specified as follows:

$$ \frac{1}{D} \sqrt{  \sum_{1\leq d \leq D} \sum_{1\leq n_d \leq N_d} \frac{SE(x_{n_d,d}-\hat{x}_{n_d, d})}{N_d} } $$

Where $D$ is the number of features, $N_d$ is the number of unobserved slots to be imputed for $d$th variable. $SE$ stands for squared errors. The results are summarized in Table \ref{tab:imputation_error}. We can see that the results are consistent with our NLL evaluations in Table \ref{tab:conditional_llh} from our main text. 

\begin{table}[h]
\setlength{\tabcolsep}{2pt}
\centering
\caption{Data imputation error on Bank dataset (averaged per variable)}
\resizebox{0.7\linewidth}{!}{
\begin{tabular}{p{0.6in}p{0.65in}p{0.65in}p{0.65in}p{0.65in}p{0.65in}p{0.65in}} \toprule
\textbf{Method} & Ours & VAE & VAE-balanced & VAE-extended & VAE-HI \\ \midrule
Bank & \textbf{0.111$\pm$0.00}& 0.116$\pm$0.00 & 0.117$\pm$0.00 &  0.116$\pm$0.00 & 0.113$\pm$0.00\\ 
Boston & \textbf{0.046$\pm$0.00} & 0.048$\pm$0.00 & 0.098$\pm$0.00 & \textbf{0.046$\pm$0.00} &0.054$\pm$0.00 \\
Avocado & \textbf{0.145$\pm$0.00}& 0.145$\pm$0.00 & 0.179$\pm$0.00& 0.145$\pm$0.00 &0.146$\pm$0.00\\ 
Energy & \textbf{0.155$\pm$0.00} & 0.176$\pm$0.00 & 0.187$\pm$0.00& 0.184$\pm$0.00 &0.176$\pm$0.00\\ 
MIMIC & \textbf{0.226$\pm$0.00}& 0.228$\pm$0.00& 0.230$\pm$0.00& 0.229$\pm$0.00 & \textbf{0.226$\pm$0.00}\\
\bottomrule
Avg. Rank & \textbf{1.00$\pm$0.00}&  2.40$\pm$0.40& 5.00$\pm$0.00& 2.60$\pm$0.67& 2.60$\pm$0.60\\
\bottomrule
\end{tabular}
}
\label{tab:imputation_error}
\end{table}

\subsection{Approximation errors of marginal VAEs}\label{app:approximation_errors_of_marginal_VAEs}

One of the main differences between our VAEM and vanilla VAEs is that we introduce one additional marginal VAE per data dimension. We evaluated the posterior approximation quality in these marginal VAEs in the Avocado dataset. The table below shows that, in each marginal VAE, the ELBO and log-likelihood are very similar:

\begin{table}[h]
\setlength{\tabcolsep}{2pt}
\centering
\caption{ELBO vs log likelihood of marginal VAEs}
\resizebox{0.7\linewidth}{!}{
\begin{tabular}{p{0.6in}p{0.5in}p{0.5in}p{0.5in}p{0.5in}p{0.5in}p{0.5in}p{0.5in}p{0.5in}} \toprule
\textbf{Variables} & 1 & 2 & 3 & 4 & 5 & 6 & 7 & 8 \\ \midrule
ELBO & -1.30 & -4.21 & -2.43& -3.49& 1.97 &2.11& 2.12& 2.17  \\ 
LLH &-1.30 & -4.17 & -2.42 & -3.48 & 2.00& 2.14 & 2.13 & 2.19  \\
\bottomrule
\end{tabular}
}
\label{tab:approximation_error}
\end{table}
\newpage
Since the gap between ELBO and LL is the KL divergence, we conclude that the posterior approximation quality is very high in this case.

In addition, our results (e.g. Figure 1 in the paper) show that our method approximates the marginal distributions of the data better than vanilla VAE.

\section{Additional Plots on Bank dataset} \label{app:exp_plots}

\begin{figure}[h]
\centering
\centering
   \includegraphics[width=17cm]{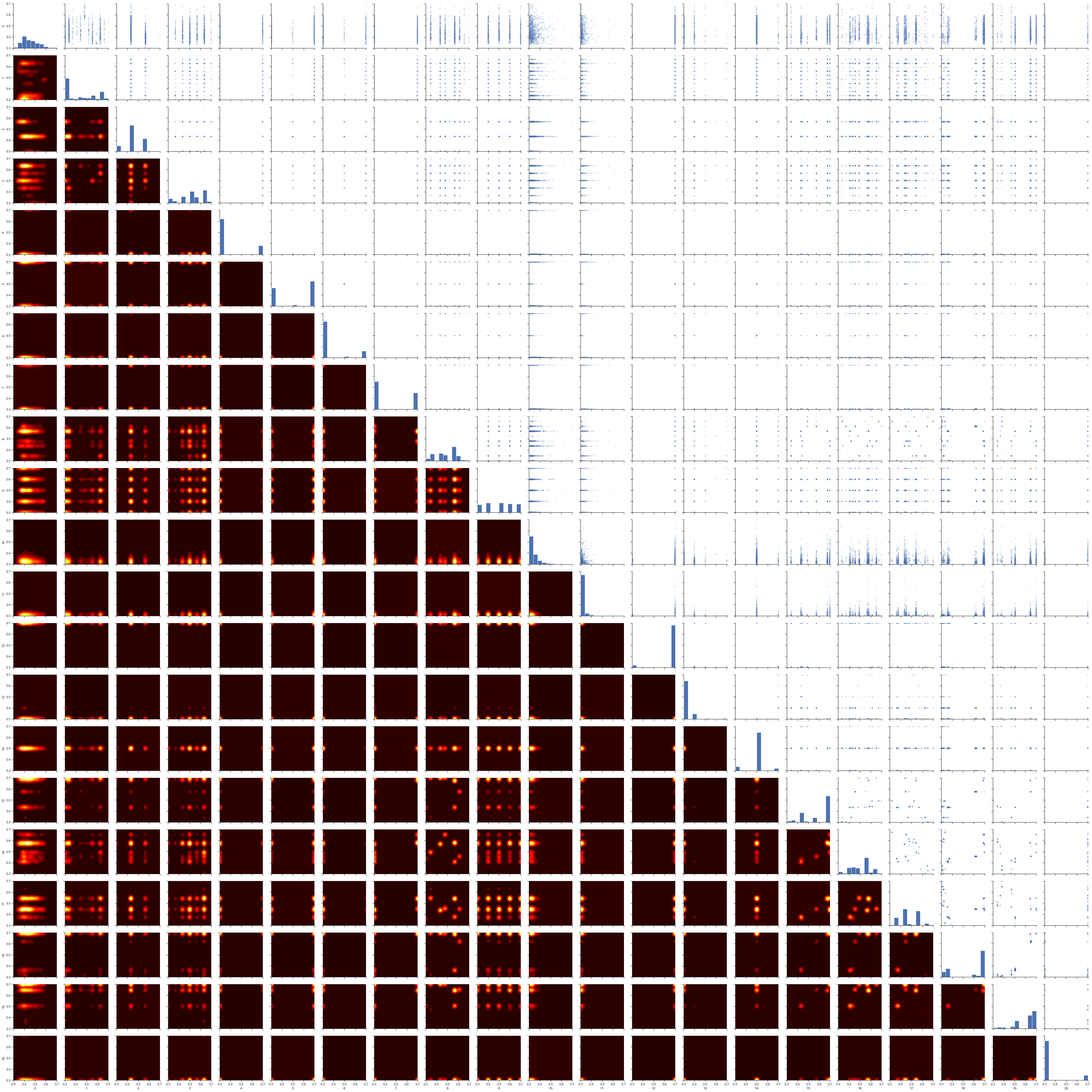}
   \caption{pair plots of all variables from the real Bank dataset. Diagonal plots show marginal histograms for each variable. The upper-triangular part shows sample scatter plots for each variable pair. The lower-triangular part shows heat maps identifying regions of high-probability density for each variable pair. For visualization, categorical variables are mapped to a grid of evenly spaced points in the interval $[0,1]$. }
\end{figure}

\begin{figure}[h]
\centering
\centering
   \includegraphics[width=17cm]{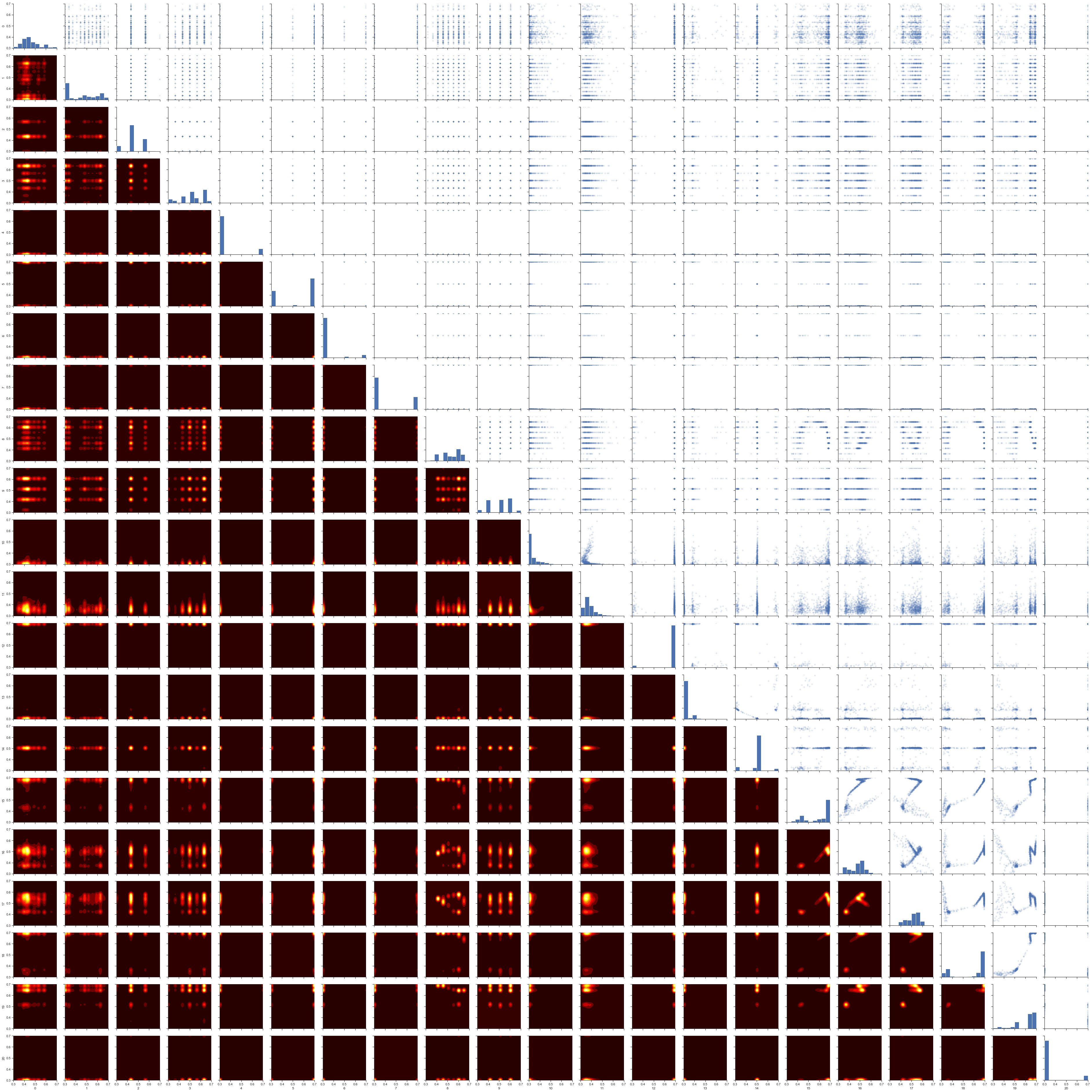}
   \caption{pair plots of all variables generated by VAEM. Diagonal plots show marginal histograms for each variable. The upper-triangular part shows sample scatter plots for each variable pair. The lower-triangular part shows heat maps identifying regions of high-probability density for each variable pair. For visualization, categorical variables are mapped to a grid of evenly spaced points in the interval $[0,1]$.   }
\end{figure}

\begin{figure}[h]
\centering
\centering
   \includegraphics[width=17cm]{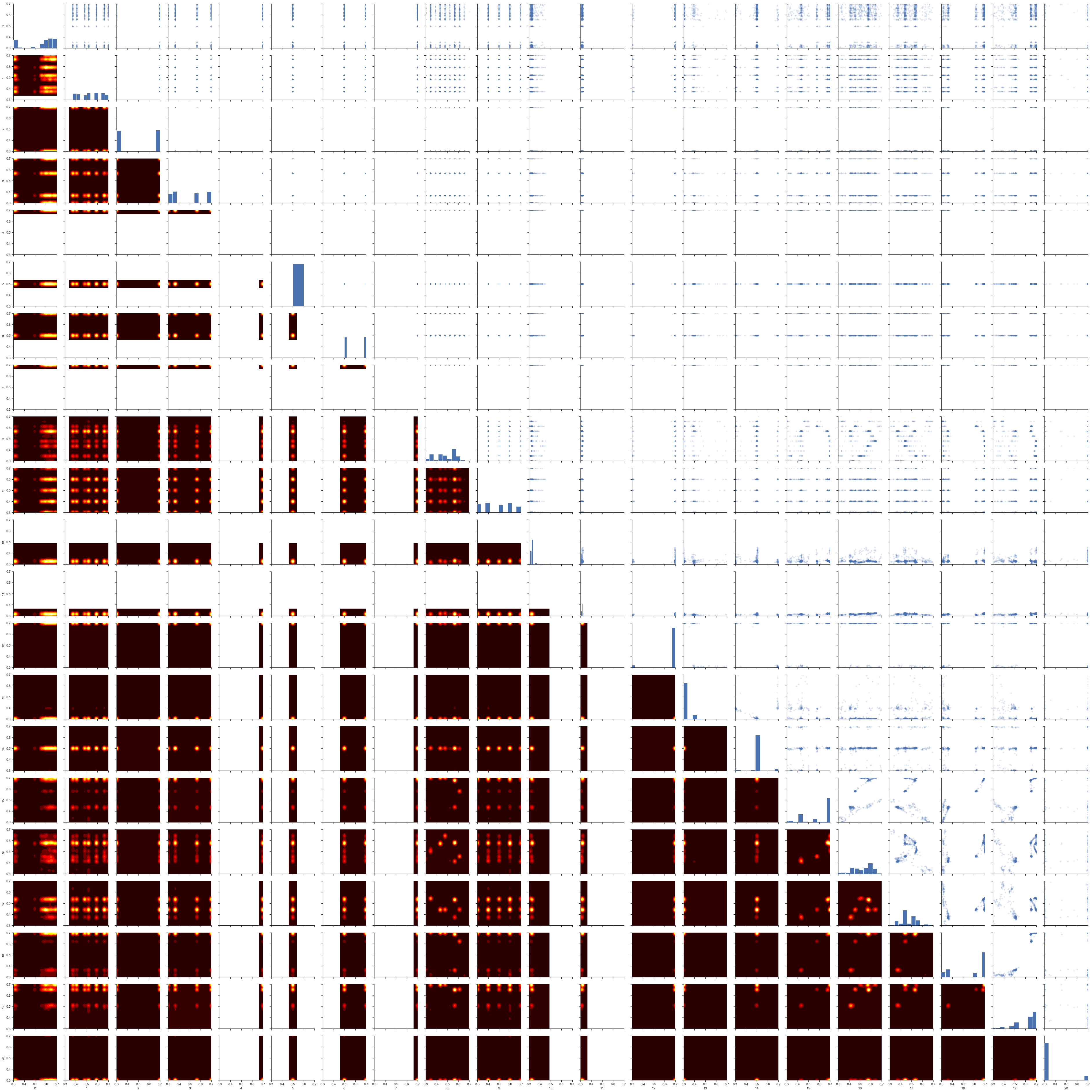}
   \caption{pair plots of all variables generated by VAE-balanced.  Diagonal plots show marginal histograms for each variable. The upper-triangular part shows sample scatter plots for each variable pair. The lower-triangular part shows heat maps identifying regions of high-probability density for each variable pair. For visualization, categorical variables are mapped to a grid of evenly spaced points in the interval $[0,1]$. }
\end{figure}

\begin{figure}[h]
\centering
\centering
   \includegraphics[width=17cm]{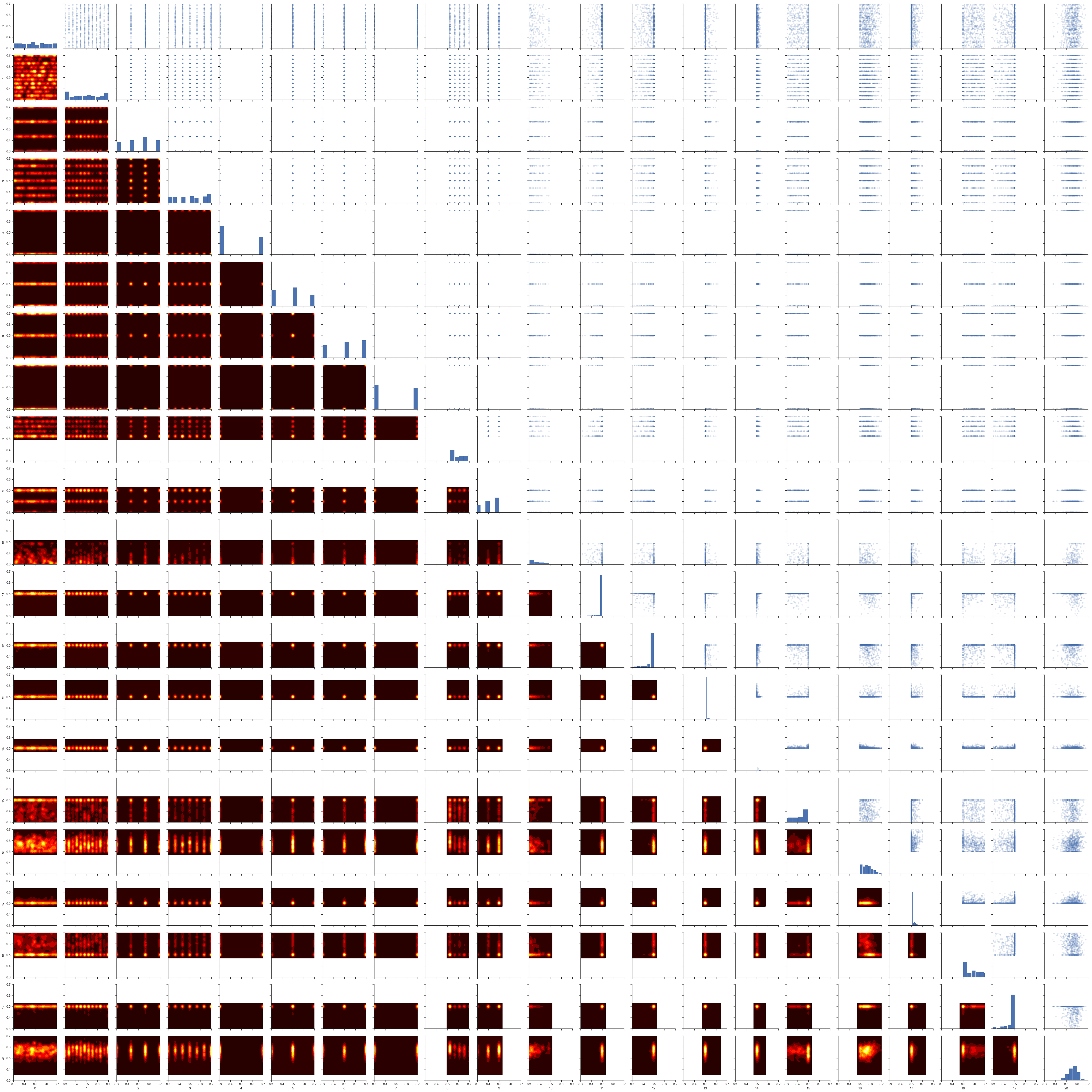}
   \caption{pair plots of all variables generated by HI-VAE.  Diagonal plots show marginal histograms for each variable. The upper-triangular part shows sample scatter plots for each variable pair. The lower-triangular part shows heat maps identifying regions of high-probability density for each variable pair. For visualization, categorical variables are mapped to a grid of evenly spaced points in the interval $[0,1]$. }
\end{figure}

\begin{figure}[h]
\centering
\centering
   \includegraphics[width=17cm]{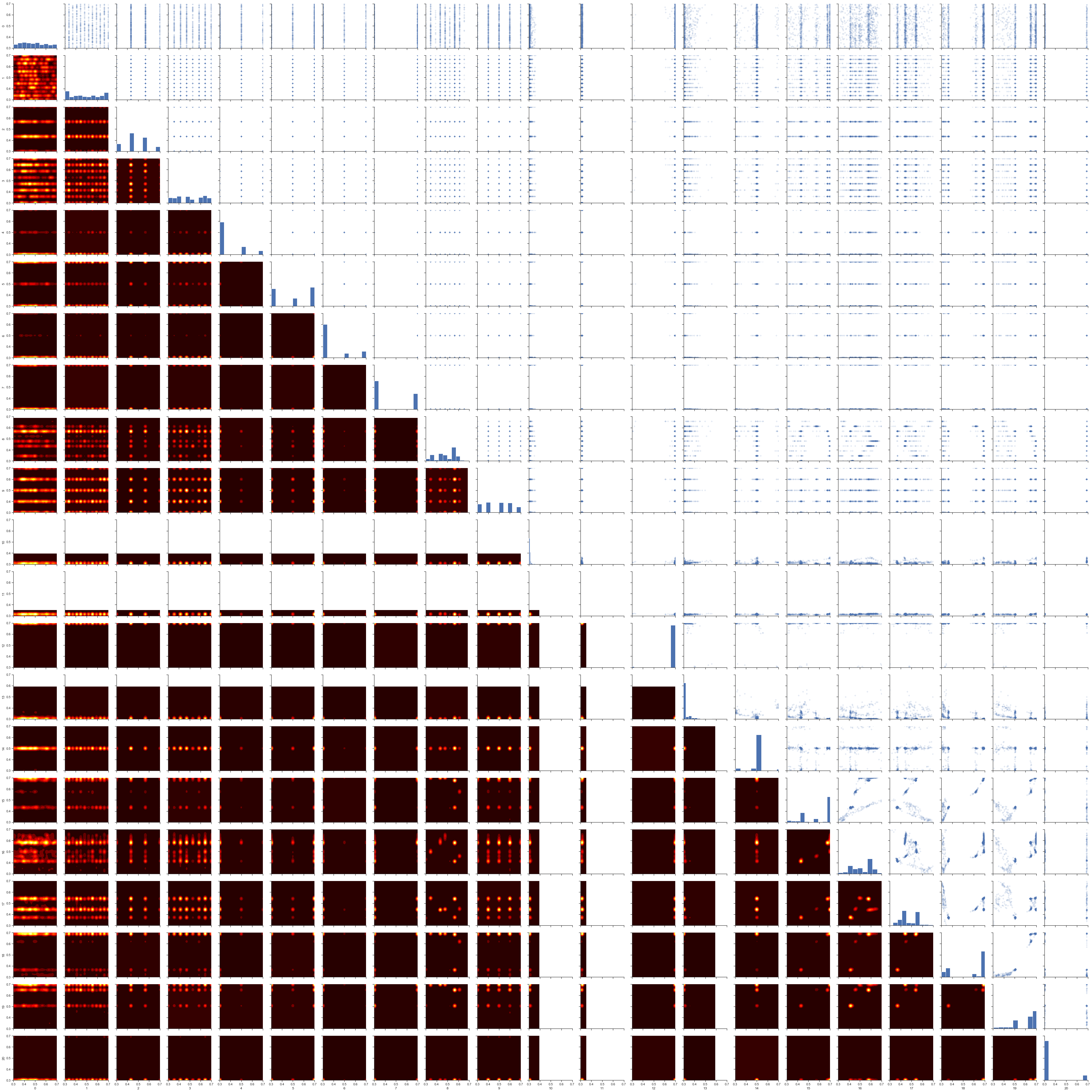}
   \caption{pair plots of all variables generated by VAE-extended.  Diagonal plots show marginal histograms for each variable. The upper-triangular part shows sample scatter plots for each variable pair. The lower-triangular part shows heat maps identifying regions of high-probability density for each variable pair. For visualization, categorical variables are mapped to a grid of evenly spaced points in the interval $[0,1]$.  }
   \label{fig:pair_VAE_extended}
\end{figure}

\end{document}